\documentclass[10pt,twocolumn,letterpaper]{article}

\usepackage{cvpr}              

\usepackage[dvipsnames]{xcolor}

\definecolor{cvprblue}{rgb}{0.21,0.49,0.74}
\usepackage[pagebackref,breaklinks,colorlinks,citecolor=cvprblue]{hyperref}
\usepackage{tabularray}
\usepackage{multirow}
\usepackage{soul}
\usepackage{caption}
\usepackage{bm}
\usepackage{makecell}
\usepackage{pifont}
\newcommand{\cmark}{\ding{51}}%
\newcommand{\xmark}{\ding{55}}%
\usepackage{float}
\usepackage{amssymb}
\usepackage{amsmath,amsthm}

\def\paperID{5761} 
\def\confName{CVPR}
\def\confYear{2024}

\title{MMM: Generative \underline{M}asked \underline{M}otion \underline{M}odel}

\author{Ekkasit Pinyoanuntapong, 
Pu Wang, 
Minwoo Lee\\
{\tt\normalsize University of North Carolina at Charlotte}\\
{\tt\small \{epinyoan, pwang13, Minwoo.Lee\}@uncc.edu}
\and
Chen Chen \\
{\tt\normalsize University of Central Florida}\\
{\tt\small chen.chen@crcv.ucf.edu}
}

\let\oldtwocolumn\twocolumn
\renewcommand\twocolumn[1][]{%
    \oldtwocolumn[{#1}{
    \begin{center}
            \includegraphics[width=\textwidth]{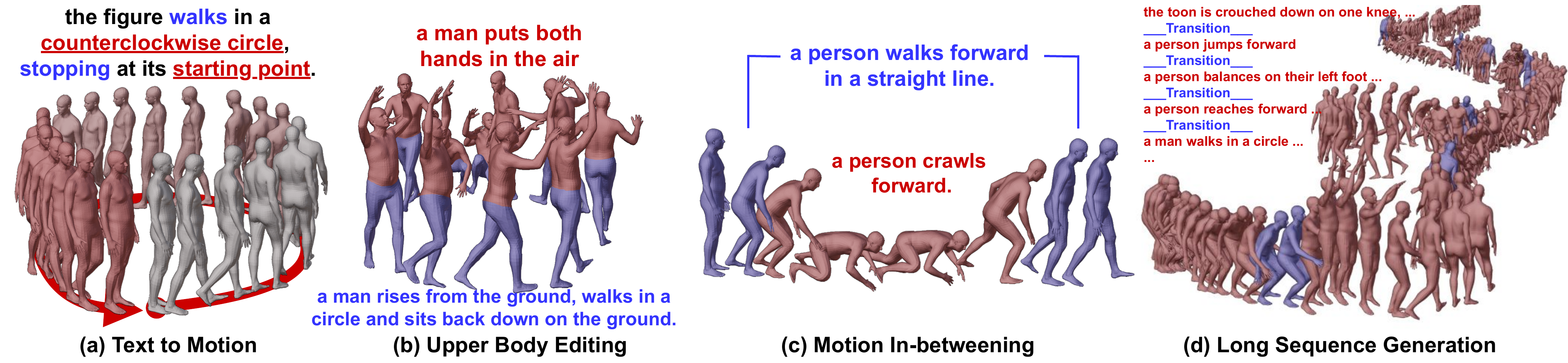}
            \vspace{-15pt}
            \captionof{figure}{MMM can generate precise human motions given fine-grained textual descriptions while enabling motion editing applications.  Blue frames represent conditioned motion input and red frames are the generated motion. Motion in-betweening can be performed by filling the gaps between keyframes or major motion points, conditioned on text or without conditions.  Upper body part editing is done by fixing the lower body motion generated by one text prompt and altering the upper body motion according to another input prompt. An arbitrary long motion sequence can be generated according to a story (i.e., a sequence of text prompts), where MMM generates the motion for each prompt (red frames),  while ``hallucinating" the nature and smooth motion transitions (blue frames) between neighboring prompts (without being explicitly trained on motion transition datasets)
            }
                  \label{fig:visualize_all_edit}
            
        \end{center}
    }]
}

\begin{document}
\maketitle



\begin{abstract}

Recent advances in text-to-motion generation using diffusion and autoregressive models have shown promising results. However, these models often suffer from a trade-off between real-time performance, high fidelity, and motion editability. To address this gap, we introduce MMM, a novel yet simple motion generation paradigm based on \textbf{M}asked \textbf{M}otion \textbf{M}odel. MMM consists of two key components: (1) a motion tokenizer that transforms 3D human motion into a sequence of discrete tokens in latent space, and (2) a conditional masked motion transformer that learns to predict randomly masked motion tokens, conditioned on the pre-computed text tokens. By attending to motion and text tokens in all directions, MMM explicitly captures inherent dependency among motion tokens and semantic mapping between motion and text tokens. During inference, this allows parallel and iterative decoding of multiple motion tokens that are highly consistent with fine-grained text descriptions, therefore simultaneously achieving high-fidelity and high-speed motion generation. In addition, MMM has innate motion editability. By simply placing mask tokens in the place that needs editing, MMM automatically fills the gaps while guaranteeing smooth transitions between editing and non-editing parts. Extensive experiments on the HumanML3D and KIT-ML datasets demonstrate that MMM surpasses current leading methods in generating high-quality motion (evidenced by superior FID scores of 0.08 and 0.429), while offering advanced editing features such as body-part modification, motion in-betweening, and the synthesis of long motion sequences. In addition, MMM is two orders of magnitude faster on a single mid-range GPU than editable motion diffusion models. Our project page is available at \url{https://exitudio.github.io/MMM-page/}.





\end{abstract}    
\section{Introduction}
\label{sec:intro}

Text-driven human motion generation has recently become an emerging research focus due to the semantic richness and user-friendly nature of natural language descriptions, with its broad applications in animation, film, VR/AR, and robotics. However, generating high-fidelity motion that precisely aligns with text descriptors is challenging because of inherent differences between language and motion data distributions. To address this challenge, three predominant methods have been proposed, including (1) language-motion latent space alignment, (2) conditional diffusion model, and (3) conditional autoregressive model. 



In the first method, text descriptions and motion sequences are projected into separate latent spaces, forcibly aligned by imposing distance loss functions such as cosine similarity and KL losses  \cite{Language2Pose, TEMOS,t2m,MotionCLIP,TMR,CrossModalRF}. Due to unavoidable latent space misalignment, this method falls short of achieving high-fidelity motion generation, which requires the synthesized motion to accurately reflect the fine-grained textural descriptions. To this end, conditional diffusion and autoregressive models are proposed recently \cite{MDM,MotionDiffuse,FLAME,Fg-T2M,DiverseMotion,T2M-GPT,AttT2M, MotionGPT}. Instead of brutally forcing latent space alignment, these models learn a probabilistic mapping from the textural descriptors to the motion sequences. However, the improvement in quality comes at the cost of motion generation speed and editability.

Motion-space diffusion models learn text-to-motion mapping by applying diffusion processes to raw motion sequences conditioned on text inputs  \cite{MDM,MotionDiffuse,FLAME,Fg-T2M}. The use of the raw motion data supports partial denoising on certain motion frames and body parts, naturally supporting semantic motion editing, such as motion inpainting and body part editing. However, the redundancy in raw data usually leads to high computational overhead and thus slow motion generation speed. A recent latent-space motion diffusion model \cite{MLD} accelerates motion generation speed by compressing raw motion data into a single latent embedding. Nonetheless, this embedding hides rich temporal-spatial semantics present in the original motion data, hindering effective motion editing. 

\begin{figure}[ht]
 \centering  
  \vspace{-5pt}
  \hspace{-10pt}
 \includegraphics[width=.48\textwidth]{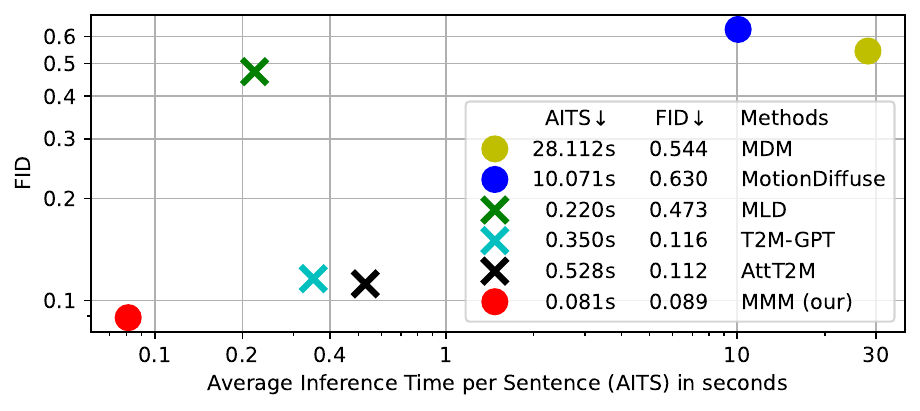}
  \caption{The motion generation quality (FID score) and speed (AITS) comparisons between MMM and SOTA methods on HumanML3D dataset. The model closer to the origin is better. MMM achieves the best FID score (0.08) and the highest speed (0.081 AITS), while preserving motion editability. ``$\bigcirc$'' represents editibility and ``$\times$'' otherwise. All tests are performed on a single NVIDIA RTX A5000.}
  \label{fig:FIDspeed}
  \vspace{-10pt}
\end{figure}

Compared with motion diffusion models, motion autoregressive models further improve motion generation fidelity by modeling temporal correlations within motion sequences \cite{T2M-GPT,AttT2M, MotionGPT}. Following a training paradigm similar to large language models such as GPT \cite{GPT}, motion autoregressive models learn to predict and generate the next motion token conditioned on the text token and previously generated motion tokens. This process, known as autoregressive decoding, contributes to improved coherence and accuracy in motion generation. However, this sequential and unidirectional decoding approach not only results in significant delays in motion generation but also makes motion editing an extremely challenging, if not impossible, task. 


As mentioned above, the existing text-driven motion generation models suffer from a trade-off between real-time performance, high fidelity, and motion editability. To address this critical issue, we introduce MMM, a novel motion generation paradigm based on conditional \textbf{M}asked \textbf{M}otion \textbf{M}odel. During the training phase, MMM follows a two-stage approach. In the first stage, a motion tokenizer is pretrained based on the vector quantized variational autoencoder (VQ-VAE) \cite{vqvae}. This tokenizer converts and quantizes raw motion data into a sequence of discrete motion tokens in latent space according to a motion codebook. A large-size codebook is learned to enable high-resolution quantization that preserves the fine-grained motion representations. In the second stage, a portion of the motion token sequence is randomly masked out, and a conditional masked transformer is trained to predict all the masked motion tokens concurrently, conditioned on both the unmasked ones and input text.

\begin{figure*}[ht]
\begin{center}
\centerline{\includegraphics[width=0.8\textwidth]{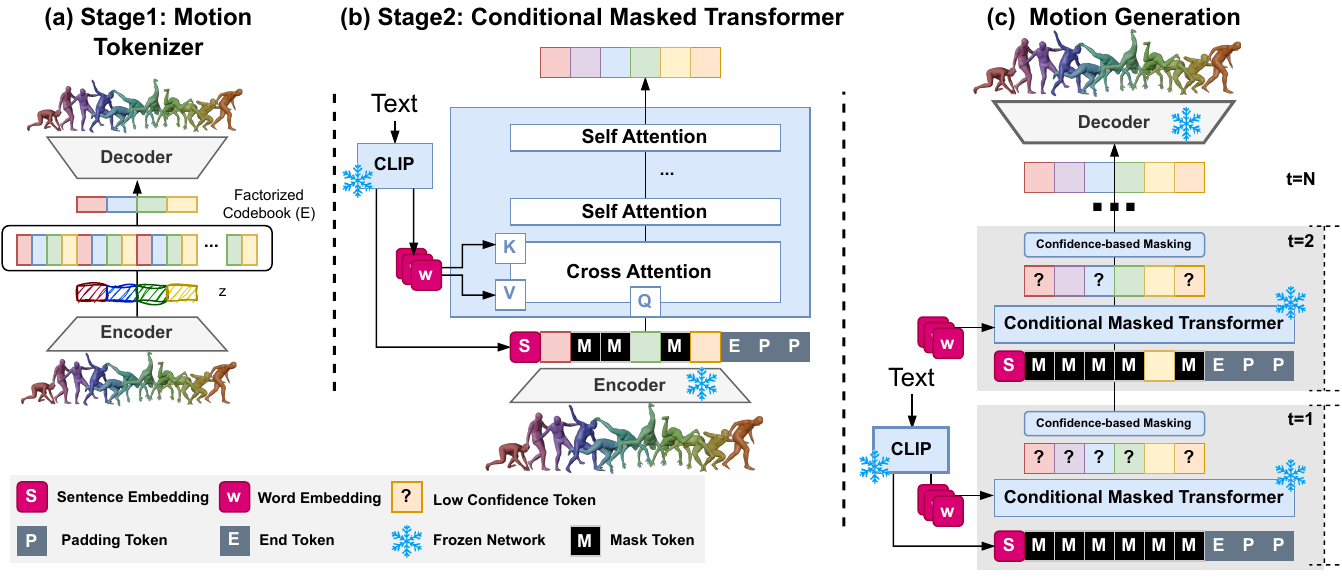}
}
\end{center}
\vspace{-30pt}
\caption{Overall architecture of MMM. \textbf{(a) Motion Tokenizer} transforms the raw motion sequence into discrete motion tokens according to a learned codebook. (b) \textbf{Conditional Masked Transformer} learns to predict masked motion tokens, conditioned on word and sentence tokens obtained from CLIP text encoders. \textbf{(c) Motion Generation} starts from an empty canvas and the masked transformer concurrently and progressively predicts multiple high-confidence motion tokens.}
\label{fig:overall_architecture}
\vspace{-10pt}
\end{figure*}


By attending to motion and text tokens in all directions, MMM explicitly captures inherent correlation among motion tokens and semantic mapping between motion and text tokens. This enables text-driven parallel decoding during inference, where in each iteration, the model concurrently and progressively predicts multiple high-quality motion tokens that are highly consistent with text descriptions and motion dynamics. This feature allows MMM to simultaneously achieve high-fidelity and high-speed motion generation. In addition, MMM has innate motion editability. By simply placing mask tokens in the place that needs editing, MMM automatically fills the gaps while ensuring smooth and natural transitions between editing and non-editing parts. Our contributions are summarized as follows.




\begin{itemize}
  \item We introduce the generative masked motion model, a novel yet simple text-to-motion generation paradigm. This model departs from diffusion and autoregressive models that currently dominate motion generation tasks.
  \item We conduct extensive qualitative and quantitative evaluations on two standard text-to-motion generation datasets, HumanML3D \cite{t2m} and KIT-ML  \cite{KIT}. It is shown in Figure \ref{fig:FIDspeed} that our model outperforms current state-of-the-art methods in both motion generation quality and speed.
  \item We demonstrate that our model enables fast and coherent motion editing via three tasks: motion in-between, upper body modification, and long sequence generation.

  
\end{itemize}

\section{Related Work}
\label{sec:relate}

\textbf{Text-driven Motion Generation.} 
Early text-to-motion generation methods are mainly based on distribution alignment between motion and language latent spaces by applying certain loss functions, such as Kullback-Leibler (KL) divergences and contrastive losses. The representative methods include Language2Pose \cite{Language2Pose}, TEMOS \cite{TEMOS}, T2M \cite{t2m}, MotionCLIP \cite{MotionCLIP}, TMR \cite{TMR} and DropTriple \cite{CrossModalRF}. Due to the significant variations between text and motion distributions, these latent space alignment approaches generally lead to unsatisfied motion generation quality.

Since denoising diffusion models \cite{DDIM}\cite{DDPM} have demonstrated notable success in vision generation tasks, \cite{GLIDE,Imagen, Make-A-Video, ImagenVideo}, diffusion models have been adopted for motion generation, where MDM \cite{MDM}, MotionDiffuse \cite{MotionDiffuse} and FRAME \cite{FLAME} are recent attempts. However, these methods directly process diffusion models in raw and redundant motion sequences, thus resulting in extremely slow during inference. In our experiments, MDM \cite{MDM} takes 28.11 seconds on average to generate a single motion on Nvidia RTX A5000 GPU. Inspired by pixel-based latent diffusion models \cite{LDM}, motion-space diffusion models MLD \cite{MLD} mitigates this issue by applying diffusion process in low-dimensional motion latent space. Nonetheless, relying on highly compressed motion embedding for speed acceleration, MLD struggles to capture fine-grained details, greatly limiting its motion editability. 

Inspired by the success of autoregressive models in language and image generations, such as GPT \cite{GPT}, DALL-E \cite{DALL-E} and VQ-GAN \cite{vqgan,Hierarchical-vqvae,vit-vqgan}, autoregressive motion models, T2M-GPT \cite{T2M-GPT}, AttT2M \cite{AttT2M} and MotionGPT \cite{MotionGPT}, have been recently developed to further improve  motion generation quality. However, these autoregressive models utilize the causal attention for unidirectional and sequential motion token prediction, limiting its ability to model bidirectional dependency in motion data, increasing the training and inference time, and hindering the motion editability.  To address these limitations,  we aim to exploit masked motion modeling for real-time, editable and high-fidelity motion generation, drawing inspiration from the success of BERT-like masked language and image modeling \cite{mask-predict,GlancingTF,BERT,M6UFCUM,UFCBERTUM,CogView2FA,MaskGIT}.



\textbf{Motion Editing.}
MDM \cite{MDM}, FLAME \cite{FLAME}, and Fg-T2M \cite{Fg-T2M} introduce text-to-motion editing, demonstrating body part editing for specific body parts and motion in-between for temporal interval adjustments. They achieve this by adapting diffusion inpainting to motion data in both spatial and temporal domains. PriorMDM \cite{PriorMDM} uses a pretrained model from MDM \cite{MDM} for three forms of motion composition: long sequence generation (DoubleTake), two-person generation (ComMDM), and fine-tuned motion control (DiffusionBlending). OmniControl \cite{OmniControl} controls any joints at any time by combining spatial and temporal control together. EDGE \cite{EDGE} introduces editing for dance generation from music tasks with similar editing capabilities, including body part editing, motion in-betweening, and long sequence generation. GMD \cite{GMD} employs the concept of editing spatial parts and temporal intervals to guide the position of the root joint (pelvic) in order to control the motion trajectory. Recently, COMODO \cite{COMODO} controls motion trajectory by integrating an RL-based control framework with the inpainting method. \ul{However, all current approaches utilize a diffusion process directly on motion space, which is slow and impractical for real-time applications.}





\section{Method}

Our goal is to design a text-to-motion synthesis paradigm that significantly improves synthesis quality, accelerates generation speed, and seamlessly preserves editability. Towards this goal, our paradigm, as depicted in Figure~\ref{fig:overall_architecture}, consists of two modules: motion tokenizer (Section~\ref{sec:pretrained_motion_tokenizer}) and conditional masked motion transformer (Section~\ref{sec:parallel-mask-motion-decoding}). Motion tokenizer learns to transform 3D human motion into a sequence of discrete motion tokens without losing rich motion semantic information. Conditional masked motion transformer is trained to predict randomly masked motion tokens conditioned on the pre-computed text tokens. During inference, masked motion transformer allows parallel decoding of multiple motion tokens simultaneously,  while considering the context from both preceding and succeeding tokens. 


\subsection{Motion Tokenizer}
\label{sec:pretrained_motion_tokenizer}
The objective of the first stage is to learn a discrete latent space by quantizing the embedding from encoder outputs $\mathbf{z}$ into the entries or codes of a learned codebook via vector quantization, as shown in Figure~\ref{fig:overall_architecture}(a). The objective of vector-quantization is defined as
\begin{equation}\label{eq:vector-quantization}
L_{V Q}=\|\operatorname{sg}(\mathbf{z})-\mathbf{e}\|_2^2+\beta\|\mathbf{z}-\operatorname{sg}(\mathbf{e})\|_2^2,
\end{equation}
where $\operatorname{sg}(\cdot ) $ is the stop-gradient operator, $\beta$ is the hyper-parameter for commitment loss, and $\mathbf{e}$ is a codebook vector from codebook $\mathbf{E}$ ($\mathbf{e} \in \mathbf{E}$). The closet Euclidean distance of the embedding $\mathbf{z}$ and codebook vector index is calculated by $i=\operatorname{argmin}_j\left\|\mathbf{z}-\mathbf{E}_j\right\|_2^2$. To preserve the fine-grained motion representations, we adopt a large codebook with a size of 8192 to reduce the information loss during embedding quantization. Our experiments show that large codebook size with a suitable code embedding dimension can lead to improved motion generation quality. However, using large codebook can aggravate codebook collapse, where the majority of tokens are assigned to just a few codes, rendering the remaining codebook inactive. To boost codebook usage,  we adopt the factorized code, which decouples code lookup and code embedding to stabilize large-size codebook learning \cite{vit-vqgan}. In addition, moving averages during codebook update and resetting dead codebooks are also adopted, which are often employed for enhancing codebook utilization in VQ-VAE and VQ-GAN \cite{Hierarchical-vqvae,vqgan}. These schemes together serve to transform 3D human motion into a sequence of discrete motion tokes in a robust and efficient manner. 


\subsection{Conditional Masked Motion Model}
\label{sec:parallel-mask-motion-decoding}


\textbf{Text-conditioned Masked Transformer.} 
During training, the motion tokens are first obtained by passing the output of the encoder through the vector quantizer. The motion token sequence, text embeddings, and special-purpose tokens serve as the inputs of a standard multi-layer transformer. Specifically, we obtain sentence embeddings from the pre-trained CLIP model \cite{CLIP} to capture the global relationships between the entire sentence and motion. The sentence embedding is then prepended to the motion tokens. Due to the nature of self-attention in transformers, all motion tokens are learned in relation to the sentence embedding. In addition, we obtain word embeddings from the same CLIP model \cite{CLIP} to capture the local relationships between each word and motion through cross-attention $=\operatorname{softmax}\left({Q_{motion} K_{\text {word}}^{T}}/{\sqrt{D}}\right) V_{\text {word}}$ \cite{AttT2M}. 

\texttt{[MASK]}, \texttt{[PAD]}, and \texttt{[END]} are learnable  special-purpose tokens. During training, the \texttt{[MASK]} token is used to represent input corruption, and the model learns to predict the actual tokens in place of \texttt{[MASK]}. During inference,  \texttt{[MASK]} tokens not only serve as placeholders for motion token generation but also prove useful in various tasks by placing the \texttt{[MASK]} tokens where editing is required, as detailed in Section~\ref{sec:motion_editing}. \texttt{[PAD]} token is used to fill up shorter motion sequences, allowing computation of batches with multiple sequences of varying lengths. Lastly, the \texttt{[END]} token is appended to the input tokens after the last token, providing the model with an explicit signal of the motion's endpoint.





\textbf{Training Strategy and Loss.}  The motion token sequence $Y$ is represented as $Y=[e_i]_{i=1}^{L}$, where $L$ denotes the sequence length. We randomly mask out $r \times L$
tokens and replace them with learnable \texttt{[MASK]} tokens, where  $r$ is the masking ratio following a uniform distribution truncated between $\alpha$ and 1. Then, the original motion token sequence $Y$ is updated with \texttt{[MASK]} tokens to form the corrupted motion sequence $Y_{\overline{\mathbf{M}}}$. This corrupted sequence along with text embedding $W$ are fed into a text-conditioned masked transformer to reconstruct input motion token sequence with reconstruction probability or confidence equal to $p\left(y_i \mid Y_{\overline{\mathbf{M}}}, W\right)$. The objective is to minimize the negative log-likelihood of the predicted masked tokens conditioned on text: 
\begin{equation}
\small
\mathcal{L}_{\text {mask}}=-\underset{\mathbf{Y} \in \mathcal{D}}{\mathbb{E}}\left[\sum_{\forall i \in[1, L]} \log p\left(y_i \mid Y_{\overline{\mathbf{M}}}, W\right)\right].
\end{equation}

\textbf{Inference via Parallel Decoding.} To decode the motion tokens during inference, we initiate the process by inputting all \texttt{[MASK]} tokens, representing an empty canvas then progressively predict more tokens per iteration. Next, iterative parallel decoding is performed, where in each iteration, the transformer masks out the subset of motion tokens that the model is least confident about and then predicts these masked tokens in parallel in the next iteration. The number of masked tokens $n_M$ is determined by a masked scheduling function, \ie a decaying function of the iteration $t$. The decaying function is chosen because at early iterations, there is high uncertainty in the predictions and thus the model begins with a large masking ratio and only keeps a small number of tokens with high prediction confidence. As the generation process proceeds, the masking ratio decreases due to the increase of context information from previous iterations. We experiment both linear $n_M = L((T-t)/T)$ \cite{mask-predict} and cosine function $n_M = L\cos(\frac{1}{2} \pi t/T)$ \cite{MaskGIT}, where cosine function yields better performance.
To obtain the final $n_M$, the length of the generated motion sequence $L$ should also be available. To address this issue, we adopt a pretrained predictor that estimates the motion sequence length based on the input text \cite{t2m}.

\section{Motion Editing}
\label{sec:motion_editing}
\begin{figure}[t]
 \centering
  \includegraphics[width=.48\textwidth]{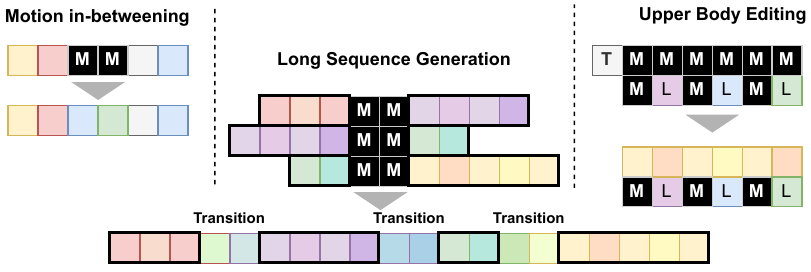}
  \vspace{-20pt}
  \caption{Motion Editing. (Left) Motion in-betweening. (Middle) Long Sequence Generation. (Right) Upper Body Editing. ``M" refers to mask token. ``T" is text conditioned tokens and ``L" denotes lower body part conditioned tokens. }
  \label{fig:motion_editing}
  \vspace{-10pt}
\end{figure}

As illustrated in Figure~\ref{fig:motion_editing},  the advantage of our masked motion modeling lies in its innate ability to edit motion.


\textbf{Motion In-betweening.}  Thanks to its mask bidirectional decoding nature, we can easily place the \texttt{[MASK]} tokens wherever editing is needed regardless of past and future context.  As an important editing task, the motion in-between involves interpolating or filling the gaps between keyframes or major motion points to create a smooth, continuous 3D animation. Since the model has already learned all possible random temporal masking combinations during training, motion in-between can be accomplished without any additional training.

\textbf{Long Sequence Generation.} Due to the limited length of motion data in the available HumanML3D \cite{t2m} and KIT \cite{KIT} datasets, where no sample exceeds a duration of 10 seconds, generating arbitrarily long motions poses a challenge. To address this, we use the trained masked motion model as a prior for long motion sequence synthesis without additional training. Particularly, given a story that consists of multiple text prompts, our model first generates the motion token sequence for each prompt. Then, we generate transition motion tokens conditioned on the end of the previous motion sequence and the start of the next motion sequence. Diffusion methods such as DoubleTakes in PriorMDM \cite{PriorMDM} require multiple steps (up to 1,000 steps) to generate transition and average the spatial differences between the previous and next motion before generating the transition motion. Our approach only needs a single step to generate realistic and natural transition motions.


\textbf{Upper Body Editing.} 
To enable body part editing, we pretrain the upper and lower body part tokenizers separately, each with its own encoders and decoders. The embedding size of each motion token is half of the regular full-body embedding size.  In the second stage, the upper and lower tokens are concatenated back to form full body embeddings. Therefore, the embedding size and the conditional masked transformer remain unchanged. Ideally, we can train the transformer by predicting the masked upper body tokens, conditioned on the text for upper body motion along with the lower body tokens. However, the generated motions are inconsistent with the text. To address this problem, we introduce random \texttt{[MASK]} tokens into lower body part motion sequence via light corruptions so that the transformer can better learn the spatial and temporal dependency of the whole body motions.  Thus, the training loss can be written as:
\begin{equation}
\small
\mathcal{L}_{\text {up}}=-\underset{\mathbf{Y} \in \mathcal{D}}{\mathbb{E}}\left[\sum_{\forall i \in[1, L]} \log p\left(y_{i}^{up} \mid Y^{up}_{\overline{\mathbf{M}}}, Y^{down}_{\overline{\mathbf{M}}}, W\right)\right]
\end{equation}
where $Y^{up}_{\overline{\mathbf{M}}}$ denotes the upper tokens with mask and $Y^{down}_{\overline{\mathbf{M}}}$ is the lower tokens with mask. It is important to note that the \texttt{[MASK]} tokens of the lower part remain unchanged throughout all iterations.



\section{Experiments}

\begin{figure*}[ht]
 \centering
  \includegraphics[width=0.95\textwidth]{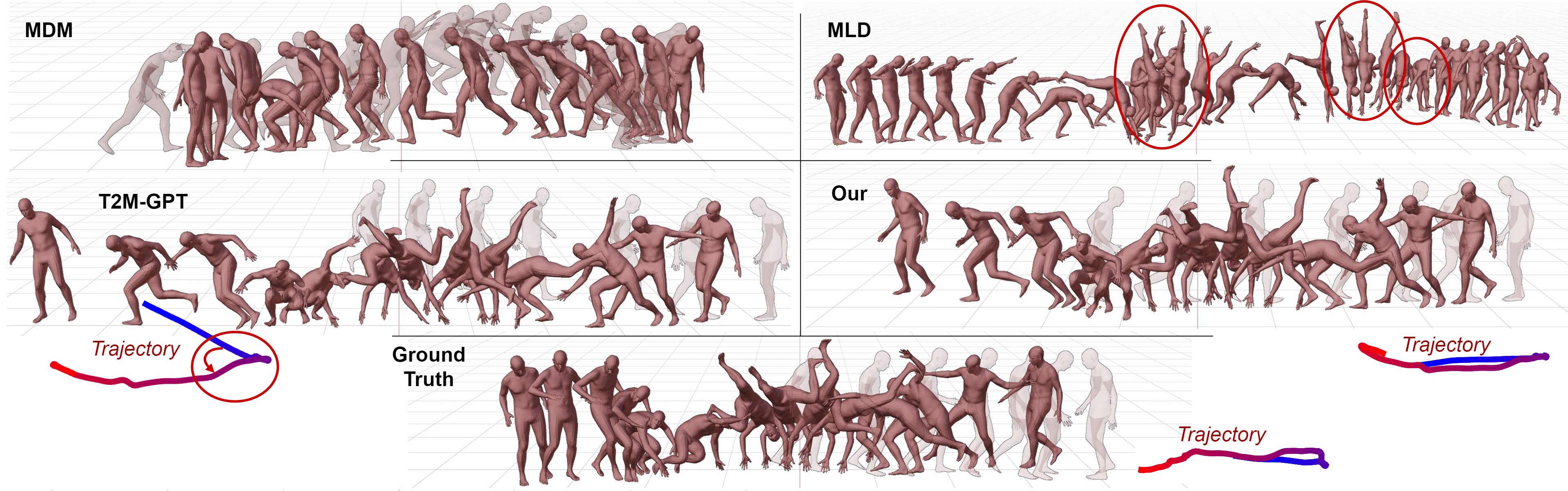}
  \vspace{-10pt}
  \caption{Qualitative comparison of state-of-the-art methods with textual description: \textbf{``a person walks forward then turns completely around and does a cartwheel.''} MDM is one of the most representative motion-space diffusion models. MLD is the first and SOTA latent-space motion diffusion model. T2M-GPT is the first and SOTA autoregressive motion model. Top-left: MDM \cite{MDM} does not execute cartwheel motion. Top-right: MLD \cite{MLD} generates unrealistic motion and lacks a complete cartwheel motion. Middle-left: the trajectory of T2M-GPT \cite{T2M-GPT} is not \textbf{``completely around"}. Middle-right: our method generates realistic motion and trajectory compared to the ground truth on the bottom. Trajectories start from \textcolor{blue}{blue} and end in \textcolor{red}{red}.}
  
  \label{fig:all_qualitative}
  \vspace{-5pt}
\end{figure*}

\begin{figure*}[ht]
 \centering
  \includegraphics[width=.8\textwidth]{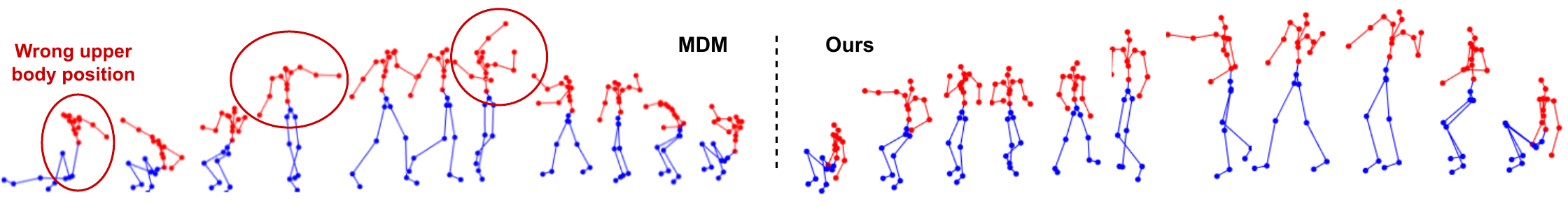}
  \caption{Qualitative comparison of \textbf{upper body editing}, generating upper body part based on the text \textcolor{red}{``a man throws a ball''} conditioned on lower body part of \textcolor{blue}{``a man rises from the ground, walks in a circle and sits back down on the ground.''}}
  \label{fig:upperbody_editting}
\end{figure*}
\begin{figure}[t]
 \centering
  \includegraphics[width=.48\textwidth]{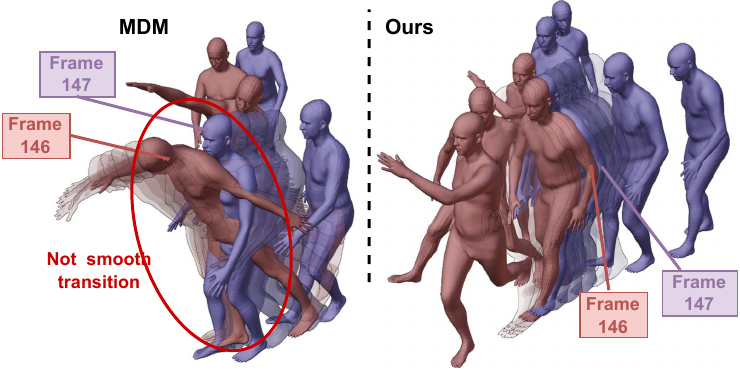}
  \caption{Qualitative comparison of \textbf{motion in-betweening}, generating 50\% motion in the middle (frame 50-146) based on the text \textcolor{red}{``a man throws a ball''} conditioned on first 25\% and last 25\% of motion of \textcolor{blue}{``a person walks backward, turns around and walks backward the other way."}. Compared with MDM, MMM achieves smoother and more natural transitions between the conditioned and generated motions (at frames 146 and 147) }
  \label{fig:motion_inbetween}
\end{figure}


\begin{table*}[ht]
\centering
\caption{\textbf{Comparison of text-conditional motion synthesis on HumanML3D \cite{t2m} test set.} For each metric, we repeat the evaluation 20 times and report the average with 95$\%$ confidence interval. The right arrow (→) indicates that the closer the result is to real motion, the better. \textcolor{red}{Red} and \textcolor{blue}{Blue} indicate the best and the second best result. $^{\S}$ reports results using ground-truth motion length.}
\vspace{-7pt}
\scalebox{0.8}{
\begin{tabular}{lccccccc} 
\hline
\multirow{2}{*}{Methods} & \multicolumn{3}{c}{R-Precision $\uparrow$}                                                                                                                                              & \multirow{2}{*}{FID $\downarrow$}                            & \multirow{2}{*}{MM-Dist $\downarrow$}                       & \multirow{2}{*}{Diversity $\rightarrow$}                    & \multirow{2}{*}{MModality $\uparrow$}        \\ 
\cline{2-4}
                         & Top-1 $\uparrow$                                            & Top-2 $\uparrow$                                            & Top-3 $\uparrow$                                            &                                                              &                                                             &                                                             &                                              \\ 
\toprule
Real & $0.511^{\pm .003}$ & $0.703^{\pm .003}$ & $0.797^{\pm .002}$ & $0.002^{\pm .000}$ & $2.974^{\pm .008}$ & $9.503^{\pm .065}$ & - \\ 
VQ-VAE & $0.503^{\pm .003}$ & $0.698^{\pm .003}$ & $0.793^{\pm .002}$ & $0.075^{\pm .001}$ & $3.027^{\pm .008}$ & $9.697^{\pm .076}$ & - \\
\toprule
Hier \cite{hier} & $0.301^{\pm .002}$ & $0.425^{\pm .002}$ & $0.552^{\pm .004}$ & $6.523^{\pm .024}$ & $5.012^{\pm .018}$ & $8.332^{\pm .042}$ & - \\
TEMOS$^{\S}$ \cite{TEMOS} & $0.424^{\pm .002}$ & $0.612^{\pm .002}$ & $0.722^{\pm .002}$ & $3.734^{\pm .028}$ & $3.703^{\pm .008}$ & $8.973^{\pm .071}$ & $0.368^{\pm .018}$ \\
TM2T \cite{TM2T} & $0.424^{\pm .003}$ & $0.618^{\pm .003}$ & $0.729^{\pm .002}$ & $1.501^{\pm .017}$ & $3.467^{\pm .011}$ & $8.589^{\pm .076}$ & $2.424^{\pm .093}$ \\
T2M \cite{t2m} & $0.455^{\pm .003}$ & $0.636^{\pm .003}$ & $0.736^{\pm .002}$ & $1.087^{\pm .021}$ & $3.347^{\pm .008}$ & $9.175^{\pm .083}$ & $2.219^{\pm .074}$ \\
MDM$^{\S}$ \cite{MDM} & $0.320^{\pm .005}$ & $0.498^{\pm .004}$ & $0.611^{\pm .007}$ & $0.544^{\pm .044}$ & $5.566^{\pm .027}$ & $\mathbf{\textcolor{red}{9.559^{\pm .086}}}$ & $\mathbf{\textcolor{blue}{2.799^{\pm .072}}}$ \\
MotionDiffuse$^{\S}$ \cite{MotionDiffuse} & $0.491^{\pm .001}$ & $0.681^{\pm .001}$ & $0.782^{\pm .001}$ & $0.630^{\pm .001}$ & $3.113^{\pm .001}$ & $9.410^{\pm .049}$ & $1.553^{\pm .042}$ \\
MLD$^{\S}$ \cite{MLD} & $0.481^{\pm .003}$ & $0.673^{\pm .003}$ & $0.772^{\pm .002}$ & $0.473^{\pm .013}$ & $3.196^{\pm .010}$ & $9.724^{\pm .082}$ & $2.413^{\pm .079}$ \\
Fg-T2M$^{\S}$ \cite{Fg-T2M} & $0.492^{\pm .002}$ & $0.683^{\pm .003}$ & $0.783^{\pm .002}$ & $0.243^{\pm .019}$ & $3.109^{\pm .007}$ & $9.278^{\pm .072}$ & $1.614^{\pm .049}$ \\
M2DM$^{\S}$ \cite{M2DM} & $0.497^{\pm .003}$ & $0.682^{\pm .002}$ & $0.763^{\pm .003}$ & $0.352^{\pm .005}$ & $3.134^{\pm .010}$ & $9.926^{\pm .073}$ & $\textcolor{red}{\mathbf{3.587^{\pm .072}}}$ \\
T2M-GPT \cite{T2M-GPT} & $0.491^{\pm .003}$ & $0.680^{\pm .003}$ & $0.775^{\pm .002}$ & $0.116^{\pm .004}$ & $3.118^{\pm .011}$ & $9.761^{\pm .081}$ & $1.856^{\pm .011}$ \\
AttT2M \cite{AttT2M} & $0.499^{\pm .003}$ & $0.690^{\pm .002}$ & $0.786^{\pm .002}$ & $0.112^{\pm .006}$ & $3.038^{\pm .007}$ & $9.700^{\pm .090}$ & $2.452^{\pm .051}$ \\
\toprule
MMM$^{\S}$ (ours) & $\textcolor{red}{\mathbf{0.515^{\pm .002}}}$ & $\textcolor{red}{\mathbf{0.708^{\pm .002}}}$ & $\textcolor{red}{\mathbf{0.804^{\pm .002}}}$ & $\textcolor{blue}{\mathbf{0.089^{\pm .005}}}$ & $\mathbf{\textcolor{red}{2.926^{\pm .007}}}$ & $\textcolor{blue}{\mathbf{9.577^{\pm .050}}}$ & $1.226^{\pm .035}$ \\


MMM (ours) & $\textcolor{blue}{\mathbf{0.504^{\pm .003}}}$ & $\textcolor{blue}{\mathbf{0.696^{\pm .003}}}$ & $\textcolor{blue}{\mathbf{0.794^{\pm .002}}}$ & $\textcolor{red}{\mathbf{0.080^{\pm .003}}}$ & $\textcolor{blue}{\mathbf{2.998^{\pm .007}}}$ & $9.411^{\pm .058}$ & $1.164^{\pm .041}$ \\

\bottomrule
\end{tabular}
}
\label{tab:humanml3d}
\end{table*}

\begin{table*}[ht]
\centering
\caption{\textbf{Comparison of text-conditional motion synthesis on KIT-ML \cite{KIT} test set.} For each metric, we repeat the evaluation 20 times and report the average with 95\% confidence interval. The right arrow (→) indicates that the closer the result is to real motion, the better. \textcolor{red}{Red} and \textcolor{blue}{Blue} indicate the best and the second best result. $^{\S}$ reports results using ground-truth motion length.}
\vspace{-7pt}
\scalebox{0.8}{
\begin{tabular}{lccccccc} 
\hline
\multirow{2}{*}{Methods} & \multicolumn{3}{c}{R-Precision $\uparrow$}                                                                                                                                              & \multirow{2}{*}{FID $\downarrow$}                            & \multirow{2}{*}{MM-Dist $\downarrow$}                       & \multirow{2}{*}{Diversity $\rightarrow$}                    & \multirow{2}{*}{MModality $\uparrow$}        \\ 
\cline{2-4}
                         & Top-1 $\uparrow$                                            & Top-2 $\uparrow$                                            & Top-3 $\uparrow$                                            &                                                              &                                                             &                                                             &                                              \\ 
\toprule
Real & $0.424^{\pm .005}$ & $0.649^{\pm .006}$ & $0.779^{\pm .006}$ & $0.031^{\pm .004}$ & $2.788^{\pm .012}$ & $11.080^{\pm .097}$ & - \\
VQ-VAE & $0.392^{\pm .006}$ & $0.606^{\pm .006}$ & $0.736^{\pm .006}$ & $0.641^{\pm .014}$ & $3.047^{\pm .012}$ & $11.075^{\pm .113}$ & - \\
\toprule
Hier \cite{hier} & $0.255^{\pm .006}$ & $0.432^{\pm .007}$ & $0.531^{\pm .007}$ & $5.203^{\pm .107}$ & $4.986^{\pm .027}$ & $9.563^{\pm .072}$ & - \\
TEMOS$^{\S}$ \cite{TEMOS} & $0.353^{\pm .006}$ & $0.561^{\pm .007}$ & $0.687^{\pm .005}$ & $3.717^{\pm .051}$ & $3.417^{\pm .019}$ & $10.84^{\pm .100}$ & $0.532^{\pm .034}$ \\
TM2T \cite{TM2T} & $0.280^{\pm .005}$ & $0.463^{\pm .006}$ & $0.587^{\pm .005}$ & $3.599^{\pm .153}$ & $4.591^{\pm .026}$ & $9.473^{\pm .117}$ & $\textcolor{blue}{\mathbf{3.292^{\pm .081}}}$ \\
T2M \cite{t2m} & $0.361^{\pm .006}$ & $0.559^{\pm .007}$ & $0.681^{\pm .007}$ & $3.022^{\pm .107}$ & $3.488^{\pm .028}$ & $10.72^{\pm .145}$ & $2.052^{\pm .107}$ \\
MDM$^{\S}$ \cite{MDM} & $0.164^{\pm .004}$ & $0.291^{\pm .004}$ & $0.396^{\pm .004}$ & $0.497^{\pm .021}$ & $9.191^{\pm .022}$ & $10.85^{\pm .109}$ & $1.907^{\pm .214}$ \\
MotionDiffuse$^{\S}$ \cite{MotionDiffuse} & $\textcolor{blue}{\mathbf{0.417^{\pm .004}}}$ & $0.621^{\pm .004}$ & $0.739^{\pm .004}$ & $1.954^{\pm .064}$ & $\textcolor{red}{\mathbf{2.958^{\pm .005}}}$ & $\textcolor{red}{\mathbf{11.10^{\pm .143}}}$ & $0.730^{\pm .013}$ \\
MLD$^{\S}$ \cite{MLD} & $0.390^{\pm .008}$ & $0.609^{\pm .008}$ & $0.734^{\pm .007}$ & $0.404^{\pm .027}$ & $3.204^{\pm .027}$ & $10.80^{\pm .117}$ & $2.192^{\pm .071}$ \\
Fg-T2M$^{\S}$ \cite{Fg-T2M} & $\textcolor{red}{\mathbf{0.418^{\pm .005}}}$ & $0.626^{\pm .004}$ & $\textcolor{blue}{\mathbf{0.745^{\pm .004}}}$ & $0.571^{\pm .047}$ & $3.114^{\pm .015}$ & $10.93^{\pm .083}$ & $1.019^{\pm .029}$ \\
M2DM$^{\S}$ \cite{M2DM} & $0.416^{\pm .004}$ & $\textcolor{blue}{\mathbf{0.628^{\pm .004}}}$ & $0.743^{\pm .004}$ & $0.515^{\pm .029}$ & $3.015^{\pm .017}$ & $11.417^{\pm .97}$ & $\textcolor{red}{\mathbf{3.325^{\pm .37}}}$ \\
T2M-GPT \cite{T2M-GPT} & $0.402^{\pm .006}$ & $0.619^{\pm .005}$ & $0.737^{\pm .006}$ & $0.717^{\pm .041}$ & $3.053^{\pm .026}$ & $10.86^{\pm .094}$ & $1.912^{\pm .036}$ \\
AttT2M \cite{AttT2M} & $0.413^{\pm .006}$ & $\textcolor{red}{\mathbf{0.632^{\pm .006}}}$ & $\textcolor{red}{\mathbf{0.751^{\pm .006}}}$ & $0.870^{\pm .039}$ & $3.039^{\pm .021}$ & $\textcolor{blue}{\mathbf{10.96^{\pm .123}}}$ & $2.281^{\pm .047}$ \\
\toprule
MMM$^{\S}$ (ours) & $0.404^{\pm .005}$ & $0.621^{\pm .005}$ & $0.744^{\pm .004}$ & $\textcolor{red}{\mathbf{0.316^{\pm .028}}}$ & $\textcolor{blue}{\mathbf{2.977^{\pm .019}}}$ & $10.910^{\pm .101}$ & $1.232^{\pm .039}$ \\

MMM (ours) & $0.381^{\pm .005}$ & $0.590^{\pm .006}$ & $0.718^{\pm .005}$ & $\textcolor{blue}{\mathbf{0.429^{\pm .019}}}$ & $3.146^{\pm .019}$ & $10.633^{\pm .097}$ & $1.105^{\pm .026}$ \\
\bottomrule
\end{tabular}
}
\label{tab:kit}
\vspace{-6pt}
\end{table*}

\begin{table}
\centering
\caption{Evaluation of two editing tasks, motion in-betweening and upper body editing, both with text and without text conditioning, in comparison to MDM \cite{MDM}, using evaluations from \cite{t2m}.}
\vspace{-7pt}
\label{tab:mask_schedule}

\scalebox{0.6}{
\begin{tblr}{
  row{3} = {c},
  row{5} = {c},
  row{7} = {c},
  row{9} = {c},
  cell{1}{2} = {c,b},
  cell{1}{3} = {c},
  cell{1}{4} = {c},
  cell{1}{5} = {c},
  cell{1}{6} = {c},
  cell{2}{1} = {r=2}{},
  cell{2}{2} = {c},
  cell{2}{3} = {c},
  cell{2}{4} = {c},
  cell{2}{5} = {c},
  cell{2}{6} = {c},
  cell{4}{1} = {r=2}{},
  cell{4}{2} = {c},
  cell{4}{3} = {c},
  cell{4}{4} = {c},
  cell{4}{5} = {c},
  cell{4}{6} = {c},
  cell{6}{1} = {r=2}{},
  cell{6}{2} = {c},
  cell{6}{3} = {c},
  cell{6}{4} = {c},
  cell{6}{5} = {c},
  cell{6}{6} = {c},
  cell{8}{1} = {r=2}{},
  cell{8}{2} = {c},
  cell{8}{3} = {c},
  cell{8}{4} = {c},
  cell{8}{5} = {c},
  cell{8}{6} = {c},
  vline{2-3} = {1-9}{},
  vline{3} = {3,5,7,9}{},
  hline{1,10} = {-}{0.08em},
  hline{2} = {1-4}{},
  hline{2} = {5-6}{0.08em},
  hline{4,6,8} = {-}{},
}
                                 & Methods      & {R-Precision \\ Top-1 $\uparrow$}        & {FID \\$\downarrow$}                    & {MM-Dist \\$\downarrow$}                   & { Diversity \\$\rightarrow$}               \\
{In-betweening\\(w/ text) }      & MDM~         & 0.389                                      & 2.371                                      & 3.859                                      & 8.077                                      \\
                                 & \textbf{Ours} & \textbf{\textbf{\textbf{\textbf{0.5226}}}} & \textbf{\textbf{\textbf{\textbf{0.0712}}}} & \textbf{\textbf{\textbf{\textbf{2.9097}}}} & \textbf{\textbf{\textbf{\textbf{9.5794}}}} \\
{In-betweening\\(w/o text) }     & MDM~ ~       & 0.284                                      & 3.417                                      & 4.941                                      & 7.472                                      \\
                                 & \textbf{Ours} & \textbf{\textbf{\textbf{\textbf{0.4239}}}} & \textbf{\textbf{0.114}}                    & \textbf{\textbf{3.6310}}                   & \textbf{\textbf{9.3817}}                   \\
{Upper Body Editing\\(w/ text)}  & MDM~         & 0.298                                      & 4.827                                      & 4.598                                      & 7.010                                      \\
                                 & \textbf{Ours} & \textbf{\textbf{0.500}}                    & \textbf{\textbf{0.1026}}                   & \textbf{\textbf{2.9720}}                   & \textbf{\textbf{9.2540}}                   \\
{Upper Body Editing\\(w/o text)} & MDM~ ~       & 0.258                                      & 7.436                                      & 5.075                                      & 6.647                                      \\
                                 & \textbf{Ours} & \textbf{\textbf{0.4834}}                   & \textbf{\textbf{0.1338}}                   & \textbf{\textbf{3.1876}}                   & \textbf{\textbf{9.0092}}                   
\end{tblr}
}
\label{tab:editing}
\vspace{-12pt}
\end{table}

In this section, we present comparisons to evaluate our models on both quality and time efficiency. In Section~\ref{sec:compare_sota}, following the standard evaluation from \cite{t2m} across multiple datasets, we observe that our method consistently outperforms the current state-of-the-art methods. Moreover, in Section~\ref{sec:inference_speed_and_editability}, evaluating with the time efficiency metric from MLD \cite{MLD}, our method exhibits shorter inference times, both on average and with respect to motion lengths.

\textbf{Datasets.}
We conduct experiments on two standard datasets for text-to-motion generation: HumanML3D \cite{t2m} and KIT Motion-Language (KIT-ML) \cite{KIT} and follow the evaluation protocol proposed in \cite{t2m}. \textbf{KIT Motion-Language (KIT-ML)} \cite{KIT} contains 3911 motion sequences and 6,278 text descriptions, with an average of 9.5 words per annotation. One to four textual descriptions are provided for each motion clip, with an average description length of around 8 sentences. Motion sequences are selected from KIT \cite{KIT} and CMU \cite{CrossModalRF} datasets but have been downsampled to a rate of 12.5 frames per second (FPS). The dataset is split into training, validation, and test sets, with respective proportions of 80\%, 5\%, and 15\%.  \textbf{HumanML3D} \cite{t2m} is currently the largest 3D human text-motion dataset, covers a diverse array of everyday human actions, including activities like exercising and dancing. The dataset comprises 14,616 motion sequences and 44,970 text descriptions. The entirety of the textual descriptions consists of 5,371 unique words. The motion sequences are originally from AMASS \cite{AMASS} and HumanAct12 \cite{HumanAct12}. Each sequence is adjusted to 20 frames per second (FPS) and trimmed to a 10-second duration, resulting in durations ranging from 2 to 10 seconds. Each motion clip is paired with at least three corresponding descriptions, with the average description length being approximately 12 words. Similar to KIT, the dataset is split into training, validation, and test sets, with respective proportions of 80\%, 5\%, and 15\%. 

\textbf{Evaluation Metrics.} The embeddings of textual descriptions and motion are encoded by pre-trained models from \cite{t2m} for evaluation metrics as proposed by \cite{t2m} with five metrics. \textbf{R-precision} and \textbf{Multimodal Distance (MM-Dist)} measure how well the generated motions align with the input prompts. Top-1, Top-2, and Top-3 accuracy are reported for R-Precision. \textbf{Frechet Inception Distance (FID)} measures the distribution distance between the generated and ground truth motion features. \textbf{Diversity} is calculated by averaging Euclidean distances of random samples from 300 pairs of motion, and \textbf{MultiModality (MModality)} represents the average variance for a single text prompt by computing Euclidean distances of 10 generated pairs of motions.

\subsection{Comparison to State-of-the-art Approaches}
\label{sec:compare_sota}

We evaluate our methods with state-of-the-art approaches \cite{hier,TEMOS,TM2T,t2m,MDM,MotionDiffuse,MLD,Fg-T2M,M2DM,T2M-GPT,AttT2M} on HumanML3D \cite{t2m} and KIT-ML \cite{KIT}. We maintain the same architecture and hyperparameters for the evaluation on both datasets.

\textbf{Quantitative comparisons.} Following \cite{t2m}, we report the average over 20 repeated generations with 95\% confidence interval. Table~\ref{tab:humanml3d} and Table~\ref{tab:kit} present evaluations on HumanML3D \cite{t2m} and KIT-ML \cite{KIT} dataset respectively,  in comparison to state-of-the-art (SOTA) approaches. Our method consistently performs best in terms of FID and Multimodal Distance. For the R-Precision and Diversity metric, our method still shows competitive results when compared to SOTA methods. On HumanML3D \cite{t2m}, as shown in Table~\ref{tab:humanml3d}, our method outperforms all SOTA methods in most of the metrics. Specifically, our method excels in Top-1, Top-2, and Top-3 R-Precision, FID, Multimodal Distance, and Diversity metrics. It is worth noting that the HumanML3D \cite{t2m} dataset is significantly larger than the KIT-ML \cite{KIT} dataset, which suggests that our method tends to perform better with more data.

\textbf{Qualitative comparison.} Figure~\ref{fig:all_qualitative} shows visual result of the generated motion by textual description, ``a person walks forward then turns completely around and does a cartwheel.'', compared to state-of-the-art models. MDM \cite{MDM} fails to generate a cartwheel motion. MLD \cite{MLD} flips in an unrealistic motion and lacks a complete cartwheel motion. Meanwhile, the trajectory of T2M-GPT \cite{T2M-GPT} does not align with the ``completely around" description in the text prompt. Our approach shows the realistic motion and trajectory compared to the ground truth. More visual results is shown in the \textbf{supplementary material}.
\subsection{Inference Speed and Editability}
\label{sec:inference_speed_and_editability}

\textbf{Inference Speed.} Figure~\ref{fig:FIDspeed} in Section~\ref{sec:intro} compares the inference speeds (AITS) of MDM \cite{MDM}, MotionDiffuse \cite{MotionDiffuse}, MLD \cite{MLD}, T2M-GPT \cite{T2M-GPT}, AttT2M \cite{AttT2M} and MMM (our method). All tests are performed on a single NVIDIA RTX A5000. It can be shown that  MMM is at least two times faster than autoregressive motion models (T2M-GPT and AttT2M) and latent-space diffusion model (MLD), while being two orders of magnitude faster than motion-space diffusion models (MDM and MotionDiffuse), which directly operate on raw motion sequences. Since the maximum motion duration of HumanML3D dataset is 9.6 seconds, AITS in Figure~\ref{fig:FIDspeed} actually measures the speed of generating short motion.  
Such high inference speed offers a significant advantage for long-range generation. As discussed in Section~\ref{sec:motion_editing}, to generate long-range motion, we combine multiple short motions with transition tokens, where the transition generation can be completed in a single iteration. In particular, our method generates a 10.873-minute sequence in only 1.658 seconds.

\textbf{Editability.} Table~\ref{tab:editing} demonstrates that our bidirectional mask modeling outperforms MDM \cite{MDM} in two editing tasks: motion in-betweening and upper body editing. We conducted experiments with and without text conditioning, evaluating on the test set of HumanML3D \cite{t2m}. Motion in-betweening is evaluated by generating 50\% of the sequences conditioned by first and last 25\% of the sequences. Qualitative results are shown in Figure~\ref{fig:visualize_all_edit}. Moreover,we also visualize the upper body editing in Figure~\ref{fig:upperbody_editting} and motion in-betweening in Figure~\ref{fig:motion_inbetween} in compared to MDM \cite{MDM}. MDM \cite{MDM} generates unrealistic upper body distinct from the lower part and exhibits discontinuous frame transitions. Our approach, on the other hand, demonstrates realistic motion and smooth transitions for editing tasks. We show more visual results in the \textbf{supplementary material}.

\section{Ablation Study}

\begin{table}
\caption{Ablation results on the masking ratio during training.}
\vspace{-7pt}
\centering
\scalebox{.74}{
\begin{tabular}{ccccccc} 
\toprule
\makecell{Masking\\Ratio}  & \makecell{R-Precision \\ Top-1 $\uparrow$} & \makecell{ FID \\ $\downarrow$}  & \makecell{MM-Dist \\ $\downarrow$} & \makecell{Diversity \\ $\rightarrow$} & \makecell{MModality \\ $\uparrow$}\\
\toprule
Uniform 0-1  & 0.513 & 0.097 & 2.923   & 9.640     & 1.097     \\
Uniform .3-1 & \textbf{0.520} & \textbf{0.089} & \textbf{2.901}   & 9.639     & 1.149     \\
Uniform .5-1  & 0.515 & \textbf{0.089} & 2.926   & \textbf{9.577}     & \textbf{1.226}     \\
Uniform .7-1  & 0.505 & 0.098 & 2.975   & 9.523     & 1.245     \\
\bottomrule
\end{tabular}
}
\label{tab:mask_schedule}
\end{table}

\begin{table}
\caption{Ablation results on speed and quality influenced by the mask scheduling during inference.}
\vspace{-7pt}
\centering
\scalebox{0.65}{
\begin{tabular}{ccccccc} 
\toprule
\makecell{\# of \\iterations}  & \makecell{AITS \\ (seconds) $\downarrow$}   & \makecell{R-Precision \\ Top-1 $\uparrow$} & \makecell{ FID \\ $\downarrow$}  & \makecell{MM-Dist \\ $\downarrow$} & \makecell{Diversity \\ $\rightarrow$} & \makecell{MModality \\ $\uparrow$}\\
\toprule
5      & \textbf{0.043} & 0.505 & 0.169 & 3.003   & 9.370     & \textbf{1.414}     \\
10     & 0.081  & 0.515 & \textbf{0.089} & 2.926   & 9.577     & 1.226     \\
15     & 0.118 & 0.516 & 0.091 & 2.919   & \textbf{9.576}     & 1.134     \\
20     & 0.149 & 0.518 & 0.096 & 2.912   & 9.590     & 1.058     \\
25     & 0.176 & 0.517 & 0.102 & 2.911   & 9.682     & 1.008     \\
30     & 0.205 & 0.515 & 0.100 & 2.908   & 9.698     & 0.937     \\
35     & 0.213 & \textbf{0.519} & 0.105 & \textbf{2.896}   & 9.701     & 0.969     \\
\bottomrule
\makecell{49\\ (Linear)} & 0.345 & 0.517 & 0.109 & 2.911   & 9.716     & 0.952     \\
\bottomrule
\end{tabular}
}
\label{tab:steps}
\end{table}

\begin{table}
\caption{Ablation results on quality influenced by the number of codes and the code dimension.}
\vspace{-7pt}
\centering
\scalebox{.80}{
\begin{tabular}{ccccccc} 
\toprule
\makecell{\# of code x \\code dimension} & \makecell{R-Precision \\ Top-1 $\uparrow$}          & \makecell{ FID \\ $\downarrow$}            & \makecell{MM-Dist \\ $\downarrow$}        & \makecell{Diversity \\ $\rightarrow$}       \\

\toprule
512 x 512               & 0.499          & 0.108          & 3.077          & 9.683           \\ 
1024 x 256              & 0.496         & 0.0850         & 3.001          & \textbf{9.641}          \\ 
2048 x 128             & 0.504          & 0.093          & 3.033          & 9.795           \\ 
4096 x 64              & \textbf{0.505} & 0.080          & 3.035          & 9.697  \\ 
8192 x 32                & 0.503          & \textbf{0.075} & \textbf{3.027} & 9.697  \\
\bottomrule
\end{tabular}
}
\label{tab:codebook_size}
\end{table}

The key success of our method lies in mask modeling. To understand how mask scheduling during training and inference impacts performance, we conduct ablation experiments using the same evaluation on HumanML3D \cite{t2m}.

\textbf{Masking Ratio during Training.} During training, we apply a random masking ratio drawn from an uniform distribution bounded between $\alpha$ and $1$. The larger $\alpha$ indicates that more aggressive masking is applied during training. As a result, the model has to predict a large number of masked tokens more based on the text prompt due to the reduced context information from unmasked motion tokens. It is shown in Table \ref{tab:mask_schedule} that too aggressive ($\alpha = 0.7$) or too mild ($\alpha = 0$) masking strategy is not beneficial for generating high-fidelity motions with sufficient diversity.



\textbf{Iteration Number.} During inference, the number of iterations directly affects the speed and the quality of the generation. The number of iterations is the maximum number of iterations for the longest motion sequence in the dataset (196 frames). As shown in Table \ref{tab:steps}, increasing the number of iterations leads to higher generation latency with slightly improved R-precision and MM-dist. However, using a small number of iterations like 10, the model already achieves the lowest FID score along with high R-precision and low MM-dist. The lowest FID score means the best overall visual quality of the generated motion, ensuring that its realism and naturalness is very close to the genuine and ground-truth human movements. High R-precision and low MM-dist indicate the precise alignment between the generated motion and the text description.



\textbf{Codebook Size.} By adopting codebook factorization, we can learn a large-size codebook for high-resolution motion embedding quantization to preserve fine-grained motion representations, which directly affect motion generation quality. In particular, codebook factorization decoupling code lookup and code embedding, where codebook uses low-dimensional latent space for code lookup and the matched code is projected back to high-dimensional embedding space.  Table \ref{tab:codebook_size} shows that motion reconstruction quality is improved by increasing the number of code entries in codebook from 512 to 8192,  while reducing the dimension of the codebook latent space from 512-d to 32-d.

\section{Conclusion}
In this work, we propose the generative masked motion model (MMM) to synthesize human motion based on textual descriptions. MMM consists of two key components: (1) a motion tokenizer converting 3D human motion into discrete latent space tokens, and (2) a conditional masked motion transformer predicting masked motion tokens based on text tokens. MMM enables parallel and iteratively-refined decoding for high-fidelity and fast motion generation. MMM has inherent motion editability. Extensive experiments demonstrate that MMM outperforms state-of-the-art methods both qualitatively and quantitatively. MMM is at least two times faster than autoregressive motion models and two orders of magnitude faster than diffusion models on raw motion sequences.
{
    \small
    \bibliographystyle{ieeenat_fullname}
    \bibliography{bib_dir/intro,bib_dir/t2m,bib_dir/diffusion,bib_dir/other,bib_dir/token}
}

\clearpage
\setcounter{page}{1}
\onecolumn
\appendix

\begin{center} 
    \centering
    \textbf{\large MMM: Generative \underline{M}asked \underline{M}otion \underline{M}odel}
\end{center}
\begin{center} 
    \centering
    \large Supplementary Material
\end{center}

\section{Overview}
\label{sec:Summary}
The supplementary material is organized into the following sections:
\begin{itemize}
    \item Section \ref{sec:sub_inferencespeed_quality}: Inference speed, quality, and editability
    \item Section \ref{sec:sub_codebookreset}: Codebook reset
    \item Section \ref{sec:sub_crossattn}: Influence of word embedding via cross attention
    \item Section \ref{sec:sub_confidence}: Confidence-based masking  
    \item Section \ref{sec:sub_inferencesRelativeLength}:  Inference speed relative to motion length 
    \item Section \ref{sec:sub_maskingschedule}: Impact of mask scheduling functions
    \item Section \ref{sec:topk-topp}: Impact of token sampling strategies
    \item Section \ref{sec:sub_qualitative_result}: Qualitative results including two new motion editing tasks: motion extrapolation/outpainting, motion completion
    \item Section \ref{sec:limitation}: Limitations
\end{itemize}
For more visualization, please visit our anonymous website \url{https://anonymous-ai-agent.github.io/MMMM} 

\section{Inference speed, quality, and editability}
\label{sec:sub_inferencespeed_quality}

To summarize the advantages of our method in three aspects—speed, quality, and editability—we compare AITS, R-Precision, FID, and Editability for MDM \cite{MDM}, MotionDiffuse \cite{MotionDiffuse}, MLD \cite{MLD}, T2M-GPT \cite{T2M-GPT}, AttT2M \cite{AttT2M} in Table \ref{tab:speed_all_models}. Only MDM \cite{MDM}, a diffusion model applied directly to motion space, and our method allows for editability. We denote this feature with a checkmark symbol, `\textcolor{red}{\cmark}'. It's worth noting that MotionDiffuse \cite{MotionDiffuse} is also capable of editability, although no specific application is provided. In the table, we represent this with a hyphen symbol, `$\boldsymbol{-}$'. While applying the diffusion process directly to motion space provides editable capabilities for various applications, the inference time of diffusion models is not practical for real-time applications. The inference time is nearly threefold the duration of the generated motion sequence. Specifically, it takes 28.112 seconds on average to generate 196 frames, which is equivalent to a 9.8-second motion sequence. On the other hand, MLD \cite{MLD} performs a diffusion process on a single motion latent space to speed up the generation time. However, the compression of its encoder not only results in the loss of fine-grained synthesis detail but also restricts its ability to edit the motion. T2M-GPT \cite{T2M-GPT} and AttT2M \cite{AttT2M} compress motion into multiple temporal embeddings, leading to faster generation time. However, due to their autoregressive approach, they lack the ability to see future tokens, which also results in a loss of editability, as indicated by the '\xmark'. Table \ref{tab:speed_all_models} shows that our approach exhibits the highest quality and preserves editable capabilities while employing only 0.081 seconds on average to generate motion.

\begin{table}[H]
\caption{Comparison of the inference speed and quality of generation on text-to-motion along with the editable capability of each model. `\textcolor{red}{\cmark}' means editable while `\xmark' is not and `$\boldsymbol{-}$' refers to has-capability but no application provided. We calculate the Average Inference Time per Sentence (AITS) on the test set of HumanML3D \cite{t2m} without model or data loading parts. All tests are performed on a single NVIDIA RTX A5000.}
\centering
\scalebox{.9}{
\begin{tabular}{ccccc} 
\toprule
Methods            & \makecell{AITS \\ (seconds)} $\downarrow$ & \makecell{R-Precision \\ Top-1 }$\uparrow$ & FID $\downarrow$ & Edit \\
\toprule
MDM                         & 28.112                                      & 0.320                                                 & 0.544                                      &  \textcolor{red}{\cmark}             \\
MotionDiffuse               & 10.071                                      & 0.491                                                 & 0.630                                      &  $\boldsymbol{-}$   \\
MLD                         & 0.220                                       & 0.481                                                 & 0.473                                      & \xmark              \\
T2M-GPT                     & 0.350                                       & 0.491                                                 & 0.116                                      & \xmark              \\
AttT2M                      & 0.528                                       & 0.499                                                 & 0.112                                      & \xmark              \\
\bottomrule
\makecell{\textbf{MMM (our)}} & $\textcolor{red}{\mathbf{0.081}}$                              & $\textcolor{red}{\mathbf{0.515}}$                                        & $\textcolor{red}{\mathbf{0.089}}$                             &  \textcolor{red}{\cmark}        \\
\bottomrule
\end{tabular}
}
\label{tab:speed_all_models}
\end{table}

\section{Codebook Reset}
\label{sec:sub_codebookreset}
\begin{figure}[H] 
\centering
\scalebox{1}{
\centerline{\includegraphics[width=\textwidth]{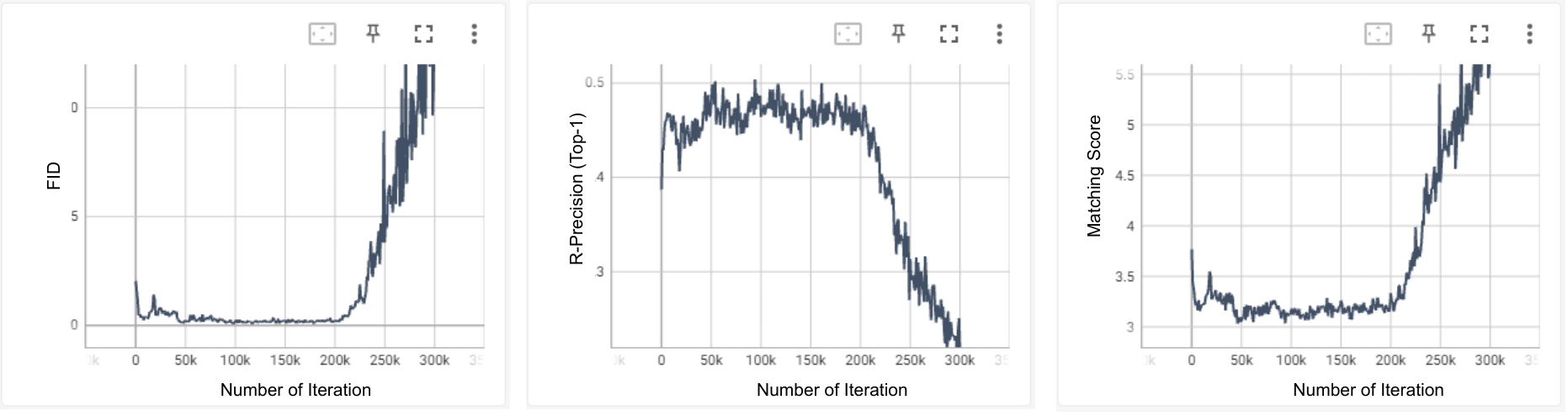}}
}
\caption{Codebook collapse effects in FID score, R-precision, and Multi-modal Distance metrics on the evaluation set during training using codebook reset in every training iteration.}
\label{fig:diverge}
\end{figure}

During stage 1, the motion tokenizer learns discrete tokens as discussed in \ref{sec:pretrained_motion_tokenizer}. Besides codebook factorization, we also adopt codebook reset to prevent codebook collapse, where the majority of tokens are allocated to only a few codes, while the rest of the codebook entries are inactive. As shown in our experiments, the codebook reset frequency directly affects codebook utilization and thus motion generation quality. For example,
simply implementing codebook reset every training iteration impedes the codebook from learning motion tokens effectively, aggravating codebook collapse. As visually depicted in Figure \ref{fig:diverge}, metrics such as FID score, R-precision, and Multi-modal Distance are initially learned during the early stages of training. However, when codebook collapse occurs, all the metrics significantly worsen. In contrast, resetting the unused codebook less frequently means the codebook will not be fully utilized, causing less fine-grain detail that the codebook can capture, which shows worse quality in 60 to 80 codebook reset iterations in Table \ref{tab:codebook_reset}. 

Moreover, the results in \textbf{Stage 1: Motion Tokenizer} and \textbf{Stage 2: Conditional Masked Transformer} may not align. This is because the Stage 1 objective is the reconstruction task without text condition, while the Stage 2 objective is to generate motion conditioned by text. Therefore, even though resetting every 40 iterations leads to the best performance in the first stage, resetting the codebook every 20 iterations works best in the text-to-motion task.

\begin{table}[H]
\centering
\caption{Ablation results on codebook reset every different number of training iterations. Codebook reset iteration in stage 2 indicates the pretrained models from stage 1.}
\scalebox{.8}{
\begin{tblr}{
  cells = {c},
  cell{1}{1} = {c=8}{},
  cell{2}{1} = {r=2}{},
  cell{2}{2} = {c=3}{},
  cell{2}{5} = {r=2}{},
  cell{2}{6} = {r=2}{},
  cell{2}{7} = {r=2}{},
  cell{2}{8} = {r=2}{},
  cell{4}{2} = {c=7}{},
  cell{9}{1} = {c=8}{},
  cell{10}{1} = {r=2}{},
  cell{10}{2} = {c=3}{},
  cell{10}{5} = {r=2}{},
  cell{10}{6} = {r=2}{},
  cell{10}{7} = {r=2}{},
  cell{10}{8} = {r=2}{},
  hline{1,2,3,4,9,10,11,12,16} = {-}{},
}
\textbf{Stage1: Motion Tokenizer}                                   &                        &                                                &       &                &                      &                         &                   \\
\makecell{ Codebook Reset Every \\Number of Training Iteration} & \makecell{R-Precision $\uparrow$} &       &       & \makecell{FID $\uparrow$}   & \makecell{MM-Dist $\downarrow$} & \makecell{Diversity $\rightarrow$} & \makecell{Loss $\downarrow$}    \\
               & Top-1       & Top-2 & Top-3 &       &         &           &         \\
1 Iteration                                                     & \textcolor{red}{Diverge} (as shown in Figure \ref{fig:diverge}) &                &                &                &                      &                         &                      \\
20 Iterations                                                 & 0.503                                          & 0.698          & 0.793          & 0.075          & 3.027                & 9.697                   & \textbf{0.05156}     \\
40 Iterations                                                 & \textbf{0.507}                                 & \textbf{0.698} & \textbf{0.793} & \textbf{0.059} & \textbf{3.013}       & \textbf{9.629}          & 0.05196              \\
60 Iterations                                                 & \textbf{0.507}                                 & \textbf{0.698} & \textbf{0.793} & 0.075          & 3.019                & 9.710                   & 0.05185              \\
80 Iterations                                                 & 0.505                                          & 0.697          & 0.793          & 0.079          & 3.022                & 9.658                   & 0.05163              \\

\textbf{Stage2: Conditional Masked Transformer}                                   &                        &                                                &       &                &                      &                         &                   \\
\makecell{ Codebook Reset Every \\Number of Training Iterations (From 1st Stage)} & \makecell{R-Precision $\uparrow$} &       &       & \makecell{FID $\uparrow$}   & \makecell{MM-Dist $\downarrow$} & \makecell{Diversity $\rightarrow$} & \makecell{MModality $\uparrow$}    \\
               & Top-1       & Top-2 & Top-3 &       &         &           &         \\
20 Iterations & \textbf{0.515} & \textbf{0.708} & \textbf{0.804} & \textbf{0.089} & \textbf{2.926} & \textbf{9.577} & \textbf{1.226} \\
40 Iterations & 0.508          & 0.702          & 0.798          & 0.108          & 2.954          & 9.645          & 1.136          \\
60 Iterations & 0.518          & 0.710          & 0.805          & 0.0923         & 2.897          & 9.7163         & 1.157          \\
80 Iterations & 0.507          & 0.701          & 0.797          & 0.111          & 2.963          & 9.541          & 1.208       
\end{tblr}}
\label{tab:codebook_reset}
\end{table}

\section{Influence of Word Embedding via Cross Attention}
\label{sec:sub_crossattn}
As discussed in Section. \ref{sec:parallel-mask-motion-decoding}, word embedding and sentence embedding help the model learn global and local text-to-motion relationship. We ablate the number of cross attention between word and motion embeddings to study the influence of word embedding. We replace self-attention with cross-attention at the starting layers while maintaining the same total number of 18 layers. The results presented in Table \ref{tab:cross_attn} indicate that a higher number of cross-attention layers leads to improved R-Precision but also worsens the FID score.



\begin{table}[H]
\centering
\caption{Ablation results on the number of cross attention layers.}
\label{tab:cross_attn}
\scalebox{.85}{
\begin{tblr}{
  cells = {c},
  cell{1}{1} = {r=2}{},
  cell{1}{2} = {c=3}{},
  cell{1}{5} = {r=2}{},
  cell{1}{6} = {r=2}{},
  cell{1}{7} = {r=2}{},
  cell{1}{8} = {r=2}{},
  hline{1,2,3,8} = {-}{},
}
{ Number of \\Cross Attention Layer} & R-Precision $\uparrow$   &                &                & FID  $\downarrow$          & MM-Dist $\downarrow$       & Diversity $\rightarrow$     & MModality $\uparrow$     \\
                                     & Top-1          & Top-2          & Top-3          &                &                &                &                \\
~0 Layer                             & 0.486          & 0.674          & 0.772          & \textbf{0.082} & 3.108          & \textbf{9.528} & 1.224          \\
~1 Layer                             & 0.515          & 0.708          & 0.804          & 0.089          & 2.926          & 9.577          & 1.226          \\
~2 Layer                             & 0.522          & 0.714          & 0.807          & 0.090          & 2.902          & 9.587          & 1.206          \\
~4 Layer                             & 0.524          & 0.716          & 0.810          & 0.094          & 2.891          & 9.651          & 1.234          \\
~9 Layer                             & \textbf{0.527} & \textbf{0.721} & \textbf{0.816} & 0.114          & \textbf{2.872} & 9.613          & \textbf{1.325} 
\end{tblr}}
\end{table}


\section{Confidence-based Masking}
\label{sec:sub_confidence}
\begin{figure}[H] 
\centerline{\includegraphics[width=.8\textwidth]{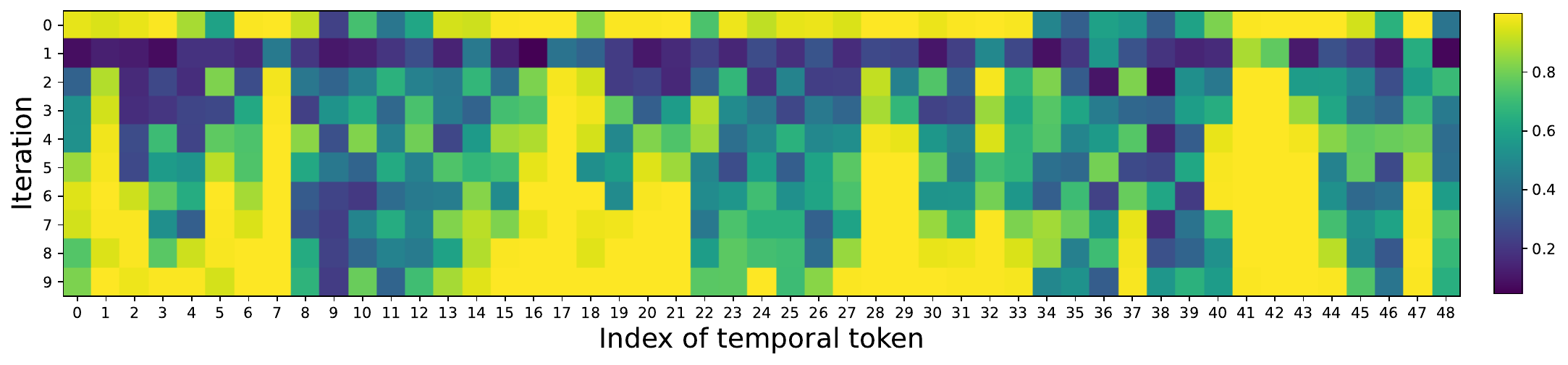}}
\caption{Visualization of the confidence in each iteration from  high confidence ($\textcolor{yellow}{\mathbf{ \blacksquare}})$ to low confidence($\textcolor{blue}{\mathbf{\blacksquare}}$)}
\label{fig:confidence}
\end{figure}

\begin{figure}[H] 
\hspace{-15pt}
\centerline{\includegraphics[width=.75\textwidth]{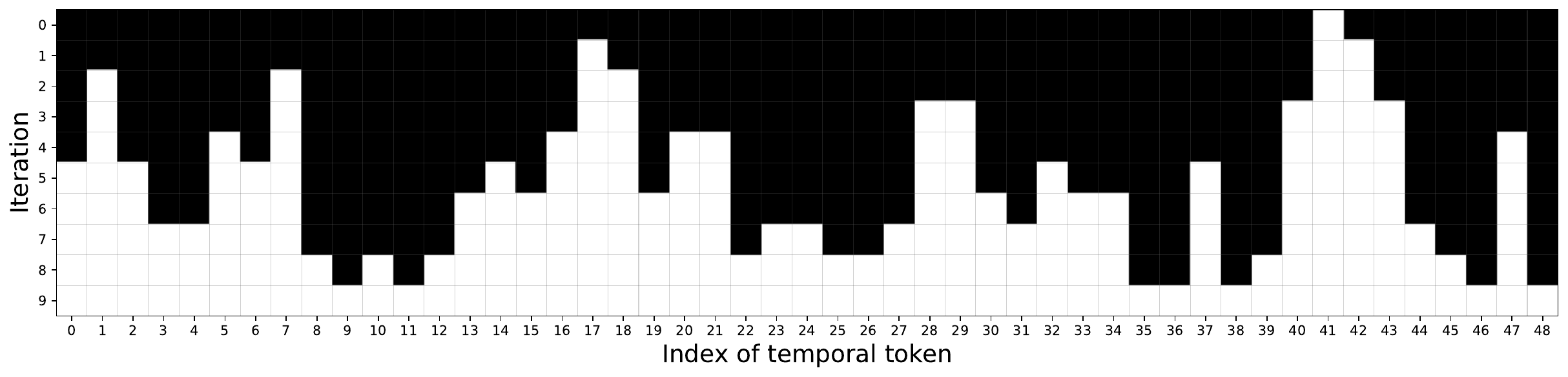}}
\caption{Visualization of mask tokens in each iteration. $\blacksquare$ indicates \texttt{[MASK]} tokens, and $\square$ refers to unmasked tokens.}
\label{fig:mask_iteration}
\end{figure}

During the inference stage, MMM adopts confidence-based masking as shown in Figure \ref{fig:overall_architecture}. To understand the behavior of this masking strategy, we visualize the confidence levels of all sequence motion tokens in each iteration. As illustrated in Figure \ref{fig:confidence} and Figure \ref{fig:mask_iteration}, the x-axis represents the indices of temporal tokens ranging from 0 to 48 (where 49 tokens correspond to 196 frames, as 4 frames are compressed into 1 token), and the y-axis indicates the 10 iterations during the generation process. The tokens with the highest confidence are predicted and used as input conditions for the next iteration. In the initial iteration (x-axis = 0), Figure \ref{fig:confidence} suggests that the model predicts most tokens with very high confidence when it solely focuses on the text condition without being constrained by other prior tokens, as all tokens are masked. However, in the second iteration, conditioned by the highest confidence token from the first iteration, the model's confidence drops before gradually increasing in the following iterations. By the final iteration, all tokens are predicted, and no \texttt{[MASK]} tokens remain, as shown in Figure \ref{fig:mask_iteration}. Notably, we observe that the increasing confidence tends to be around the location of the previous high-confidence token and expands in the later iteration.

\section{Inference Speed Relative to Motion Length}
\label{sec:sub_inferencesRelativeLength}

One notable advantage of our approach resides in its superior inference speed relative to the length of the generated motion. In particular, we use a dynamic masking schedule to calculate the number of \texttt{[MASK]} tokens $m_t$ at iteration $t$ w.r.t. maximum iteration $T$, the length of the motion tokens $L$, and the maximum number of motion tokens $M$ that can be taken by the conditional transformer as the inputs, where $M$ is determined by the maximum-length motion in the datasets. We compute the number of mask $n_M$ at $t$ iteration by a mask scheduling function  as follows:
\begin{equation}\label{scheduling}
n_M (t) =\left\lceil\gamma\left(\frac{t}{T_{dyn}}\right) \cdot L \right\rceil.
\end{equation}
where the total iteration of each sample, i.e., $T_{dyn} = T \cdot \frac{L}{M}$ with $L \leq M$, is proportional to its sample length $L$. $\gamma()$ is a decaying function. The choice of the decaying function is discussed in Section \ref{sec:sub_maskingschedule}. 

\begin{figure}[H]
\begin{center}
\scalebox{.8}{
\centerline{\includegraphics[width=.5\textwidth]{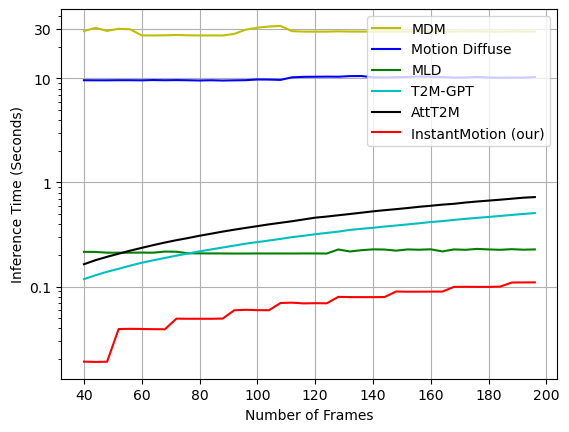}}
}
\end{center}
\caption{Comparison of Average Inference Time per Sentence (AITS) by the length of motion sequence (number of frames).}
\label{fig:speed_by_length}
\end{figure}

As shown in Figure \ref{fig:speed_by_length}, diffusion-based approaches, MDM \cite{MDM}, MotionDiffuse \cite{MotionDiffuse}, and MLD \cite{MLD}, exhibit the same inference time regardless of the number of generated frames. In contrast, token-based approaches such as T2M-GPT \cite{T2M-GPT} and AttT2M \cite{AttT2M} show slower inference speeds than MLD \cite{MLD} for long-length sequences but faster speeds for short sequences. Our approach, on the other hand, is not only faster than all state-of-the-art approaches for long sequences but even faster for short sequence generation. Specifically, the inference speed for a 40-frame sequence can be as fast as 0.018 seconds. 

This behavior offers a significant advantage for long-range generation. As discussed in Section \ref{sec:motion_editing}, to generate long-range motion, we combine multiple short motions with transition tokens, which can be generated in parallel. Since the transitions are short and our method's inference speed is relative to the motion length, transition generation can be completed in just a single iteration, as opposed to the thousand steps required by diffusion-based models. In particular, we can generate a 10.873-minute sequence in only 1.65 seconds.

\begin{figure}[H]
\centerline{\includegraphics[width=.4\textwidth]{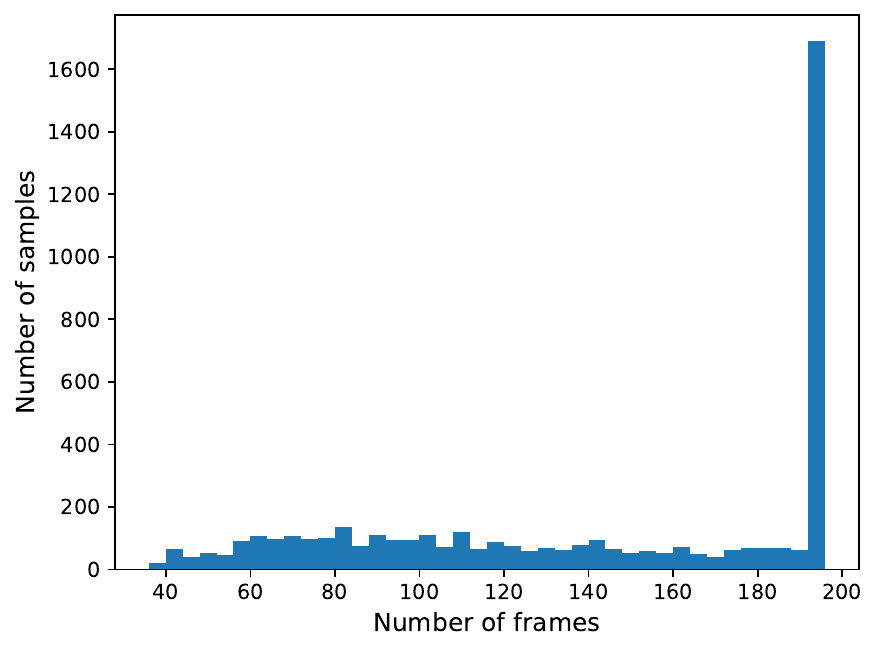}}
\caption{The number of samples in HumanML3D \cite{t2m} test set by length}
\label{fig:num_sample_by_length}
\end{figure}

It is worth noting that the HumanML3D \cite{t2m} dataset is heavily skewed towards 196-frame samples (around 1680 samples), while all other shorter samples have less than 200 samples per length. Since our method's speed is relative to the length of the generated samples, the reported average speed in Table \ref{tab:speed_all_models} could be even significantly lower if the number of samples by length was equal.  The distribution of samples by their length is visualized in Figure \ref{fig:num_sample_by_length}.

\section{Mask Scheduling Function}
\label{sec:sub_maskingschedule}
As discussed in the previous section, the mask scheduling function aims to determine the number of tokens to be masked during each iteration during inference. We experiment with three mask scheduling functions to study their influence on motion generation, varying the number of iterations. Let $L$ denote the length of generated motion and The three mask scheduling functions are described as follows:
\begin{enumerate}
    \item \textbf{Cosine} function represents a concave dependency which can be written as $n_M (t) = L \cdot cos(\frac{1}{2}\pi \frac{t}{T_{dyn}})$
    \item \textbf{Linear} function is simply an inverse relationship function of x and y: $n_M (t) =L\cdot \frac{T_{dyn} - t}{T_{dyn}}$
    \item \textbf{Square Root} function is a representative convex function, expressed as $n_M (t) = L\cdot (1-(\frac{t}{T_{dyn}})^2)$
\end{enumerate}


In Table \ref{tab:inference_schedule}, we experiment with each mask scheduling function with 5, 10, 15, 20, 25, and 30 inference iterations. The results show that \textbf{Square Root} yields the worst result. \textbf{Linear} performs the best FID score at 15 iterations and can reach the best Top-1 R-precision of 0.519. While \textbf{Cosine} can achieve its best FID score of 0.089 at only 10 iterations and reach the best Top-1 R-precision at 0.518.


\begin{table}[H]
\centering
\caption{Ablation results on different numbers of inference iterations with cosine, linear, and square root mask scheduling functions.}
\scalebox{.7}{
\begin{tblr}{
  cells = {c},
  cell{1}{1} = {r=2}{},
  cell{1}{2} = {c=3}{},
  cell{1}{5} = {r=2}{},
  cell{1}{6} = {r=2}{},
  cell{1}{7} = {r=2}{},
  cell{1}{8} = {r=2}{},
  cell{3}{1} = {c=8}{},
  cell{10}{1} = {c=8}{},
  cell{17}{1} = {c=8}{},
  hline{1,2,3,4,10,11,17,18,24} = {-}{},
}

\makecell{ Number of Inference Iterations} & \makecell{R-Precision $\uparrow$} &       &       & \makecell{FID $\uparrow$}   & \makecell{MM-Dist $\downarrow$} & \makecell{Diversity $\rightarrow$} & \makecell{MModality $\uparrow$}    \\
               & Top-1       & Top-2 & Top-3 &       &         &           &         \\
\textbf{Cosine}      &                &                &                &                &                &                &                \\
5               & 0.505          & 0.695          & 0.791          & 0.169          & 3.003          & 9.370          & \textbf{1.414} \\
10              & 0.515          & 0.708          & 0.804          & \textbf{0.089} & 2.926          & 9.577          & 1.226          \\
15              & 0.516          & 0.710          & 0.806          & 0.091          & 2.919          & \textbf{9.576} & 1.134          \\
20              & \textbf{0.518}          & 0.710          & 0.807          & 0.096          & 2.912          & 9.590          & 1.058          \\
25              & 0.517          & 0.710          & 0.805          & 0.102          & 2.911          & 9.682          & 1.008          \\
30              & 0.515 & \textbf{0.711} & \textbf{0.807} & 0.100          & \textbf{2.908} & 9.698          & 0.937          \\
\textbf{Linear~}     &                &                &                &                &                &                &                \\

5  & 0.505          & 0.693          & 0.789          & 0.195          & 2.995          & 9.399          & \textbf{1.435} \\
10 & 0.514          & 0.707          & 0.804          & 0.090 & 2.920          & 9.584          & 1.190          \\
15 & 0.518          & 0.710          & 0.805          & \textbf{0.086}          & 2.910          & \textbf{9.583} & 1.133          \\
20 & 0.519          & 0.713          & 0.808          & 0.091          & \textbf{2.900} & 9.605          & 1.057          \\
25 & 0.517          & 0.713          & 0.808          & 0.096          & \textbf{2.900} & 9.715          & 1.023          \\
30 & \textbf{0.520} & \textbf{0.713} & \textbf{0.808} & 0.103          & 2.907          & 9.632          & 0.9774     \\
\textbf{Square Root} &                &                &                &                &                &                &                \\
5  & 0.497          & 0.685          & 0.781          & 0.379          & 3.057          & 9.271          & \textbf{1.511} \\
10 & 0.508          & 0.701          & 0.798          & 0.174          & 2.960          & 9.399          & 1.297          \\
15 & 0.512          & 0.705          & 0.801          & 0.119          & 2.933          & 9.448          & 1.241          \\
20 & 0.516          & 0.708          & 0.804          & 0.099          & 2.916          & 9.582          & 1.174          \\
25 & 0.518          & 0.709          & 0.805          & \textbf{0.092} & \textbf{2.909} & \textbf{9.540} & 1.185          \\
30 & \textbf{0.520} & \textbf{0.712} & \textbf{0.807} & 0.093          & 2.910          & 9.610          & 0.9815                  
\end{tblr}}
\label{tab:inference_schedule}
\end{table}

\section{Token Sampling Strategies for Parallel Decoding}
\label{sec:topk-topp}
After the tokens are masked out during each iteration, the masked tokens are decoded in parallel in the next iteration. The decoding process is based on stochastic sampling, where the tokens are sampled based on their prediction confidences. In particular, we investigate three sampling strategies, including temperature, top-k, and top-p, on the motion generation performance. These sampling strategies are widely adopted by natural language generation tasks, where top-p sampling generally shows superior performance compared with top-k and temperature-based sampling. However, our experiments show that the temperature-based and top-k sampling yields the best performance for the motion generation task.

\textbf{Temperature Sampling.}
Temperature sampling generates the tokens according to the softmax distribution function shaped by a temperature parameter $\beta$, as shown below:
\[p({y_i}|{Y_{\bar M}},W) = \frac{{\exp ({e_i}/\beta)}}{{\sum\nolimits_{i \in E} {\exp ({e_i}/\beta)} }}\]
where $e_i$ is logits for motion code $y_i$ from the codebook and $E$ is the motion codebook. Low temperature gives more weight to the motion tokens with high prediction confidence. High-temperature trends to sample the tokens with equal probability.  Table \ref{tab:temperature} shows that with temperature $\beta = 1$, we can achieve the best overall generation performance measured by FID, while maintaining competitive performance for other metrics. 
\begin{table}[H]
\centering
\caption{Ablation results on different temperatures.}
\scalebox{.7}{
\begin{tblr}{
  cells = {c},
  cell{1}{1} = {r=2}{},
  cell{1}{2} = {c=3}{},
  cell{1}{5} = {r=2}{},
  cell{1}{6} = {r=2}{},
  cell{1}{7} = {r=2}{},
  cell{1}{8} = {r=2}{},
  hline{1,2,3,7} = {-}{},
}

\makecell{ Temperature ($\beta$)} & \makecell{R-Precision $\uparrow$} &       &       & \makecell{FID $\uparrow$}   & \makecell{MM-Dist $\downarrow$} & \makecell{Diversity $\rightarrow$} & \makecell{MModality $\uparrow$}    \\
               & Top-1       & Top-2 & Top-3 &       &         &           &         \\

.5  & \textbf{0.517} & \textbf{0.714} & \textbf{0.810} & 0.098          & \textbf{2.893} & 9.725          & 0.744          \\
1   & 0.515          & 0.708          & 0.804          & \textbf{0.089} & 2.926          & 9.577          & 1.226          \\
1.2 & 0.504          & 0.695          & 0.792          & 0.140          & 2.998          & \textbf{9.484} & 1.446          \\
1.5 & 0.478          & 0.664          & 0.761          & 0.495          & 3.186          & 9.211          & \textbf{1.884} 
\end{tblr}}
\label{tab:temperature}
\end{table}

\textbf{Top-k Sampling.}
Top-k sampling samples the token from the top k most probable choices. In particular, it first finds the top-k codebook entries to form a new codebook consisting of k entries, $E^{k} \in E$, which maximizes $\sum\nolimits_{i \in {E^k}} {p({y_i}|{Y_{\bar M}},W)}$. Then, the original distribution is re-scaled to a new distribution $p'({y_i}|{Y_{\bar M}},W)$, from which the motion token is sampled. In particular, 
$p'({y_i}|{Y_{\bar M}},W) = p({y_i}|{Y_{\bar M}},W)/{p_{sum}}$, where ${p_{sum}} = \sum\nolimits_{i \in {E^k}} {p({y_i}|{Y_{\bar M}},W)}$. ``10\%'' refers to the sampling from only the top 10\% of most probable motion tokens from the codebook entries.``100\%'' indicates the use of all codebook entries. As shown in Table \ref{tab:topk}, $k = 100\%$ leads to the best performance, which is mathematically equivalent to temperature sampling with temperature $\beta = 1$.

\begin{table}[H]
\centering
\caption{Ablation results on different probability of Top-K sampling.}
\scalebox{.7}{
\begin{tblr}{
  cells = {c},
  cell{1}{1} = {r=2}{},
  cell{1}{2} = {c=3}{},
  cell{1}{5} = {r=2}{},
  cell{1}{6} = {r=2}{},
  cell{1}{7} = {r=2}{},
  cell{1}{8} = {r=2}{},
  hline{1,2,3,9} = {-}{},
}

\makecell{ Top-k \% } & \makecell{R-Precision $\uparrow$} &       &       & \makecell{FID $\uparrow$}   & \makecell{MM-Dist $\downarrow$} & \makecell{Diversity $\rightarrow$} & \makecell{MModality $\uparrow$}    \\
               & Top-1       & Top-2 & Top-3 &       &         &           &         \\

10\% & 0.511          & 0.705          & 0.803          & \textbf{0.088} & 2.930          & 9.636          & 1.180          \\
30\% & 0.512          & 0.706          & 0.803          & 0.093          & 2.928          & 9.624          & 1.192          \\
50\% & 0.515          & 0.707          & 0.804          & 0.093          & \textbf{2.926} & 9.637          & 1.208          \\
70\% & 0.512          & 0.705          & 0.802          & 0.092          & 2.934          & \textbf{9.506} & 1.202          \\
90\% & 0.514          & 0.706          & 0.803          & 0.091          & 2.927          & 9.594          & 1.159          \\
100\%   & \textbf{0.515} & \textbf{0.708} & \textbf{0.804} & 0.089          & \textbf{2.926} & 9.577          & \textbf{1.226}  
\end{tblr}}
\label{tab:topk}
\end{table}

\textbf{Top-p Sampling.}
The Top-P strategy, also called nucleus sampling, selects the highest probability tokens whose cumulative probability mass exceeds the pre-chosen threshold $p$. In particular, it first finds the top-p codebook $E^{p} \in E$, which is the smallest set such that $\sum\nolimits_{i \in {E^k}} {p({y_i}|{Y_{\bar M}},W)} > p$.  Then, the original distribution is re-scaled to a new distribution $p'({y_i}|{Y_{\bar M}},W)$, from which the motion token is sampled. In particular, 
$p'({y_i}|{Y_{\bar M}},W) = p({y_i}|{Y_{\bar M}},W)/{p_{sum}}$, where ${p_{sum}} = \sum\nolimits_{i \in {E^{p}}} {p({y_i}|{Y_{\bar M}},W)}$. $p = 0.1$ refers to the sampling from only a set of probable tokens from the codebook entries whose summation of prediction confidences is larger than 0.1.  $p = 1$ indicates the use of all codebook entries. As shown in Table \ref{tab:topp}, selecting only 0.1 of codebook entries can improve FID score, however, worsens R-precision, MM-Dist, and MModality.

\begin{table}[H]
\centering
\caption{Ablation results on different probability of Top-P sampling.}
\scalebox{.7}{
\begin{tblr}{
  cells = {c},
  cell{1}{1} = {r=2}{},
  cell{1}{2} = {c=3}{},
  cell{1}{5} = {r=2}{},
  cell{1}{6} = {r=2}{},
  cell{1}{7} = {r=2}{},
  cell{1}{8} = {r=2}{},
  hline{1,2,3,9} = {-}{},
}

\makecell{ Probability of Top-P Sampling} & \makecell{R-Precision $\uparrow$} &       &       & \makecell{FID $\uparrow$}   & \makecell{MM-Dist $\downarrow$} & \makecell{Diversity $\rightarrow$} & \makecell{MModality $\uparrow$}    \\
               & Top-1       & Top-2 & Top-3 &       &         &           &         \\

0.1 & 0.511          & 0.704          & 0.802          & \textbf{0.084} & 2.932          & \textbf{9.518} & 0.060          \\
0.3 & 0.516          & 0.711          & 0.807          & 0.093          & 2.909          & 9.589          & 0.480          \\
0.5 & 0.516          & 0.711          & 0.807          & 0.091          & 2.905          & 9.622          & 0.739          \\
0.7 & \textbf{0.518} & \textbf{0.712} & \textbf{0.809} & 0.090          & \textbf{2.900} & 9.645          & 0.963          \\
0.9 & 0.516          & 0.710          & 0.806          & 0.089          & 2.910          & 9.636          & 1.079          \\
1   & 0.515          & 0.708          & 0.804          & 0.089          & 2.926          & 9.577          & \textbf{1.226} 
\end{tblr}}
\label{tab:topp}
\end{table}

\section{Qualitative Results}
\label{sec:sub_qualitative_result}
\subsection{Temporal Motion Editing.} 
\begin{figure}[H] 
\centering
\scalebox{.7}{
\centerline{\includegraphics[width=\textwidth]{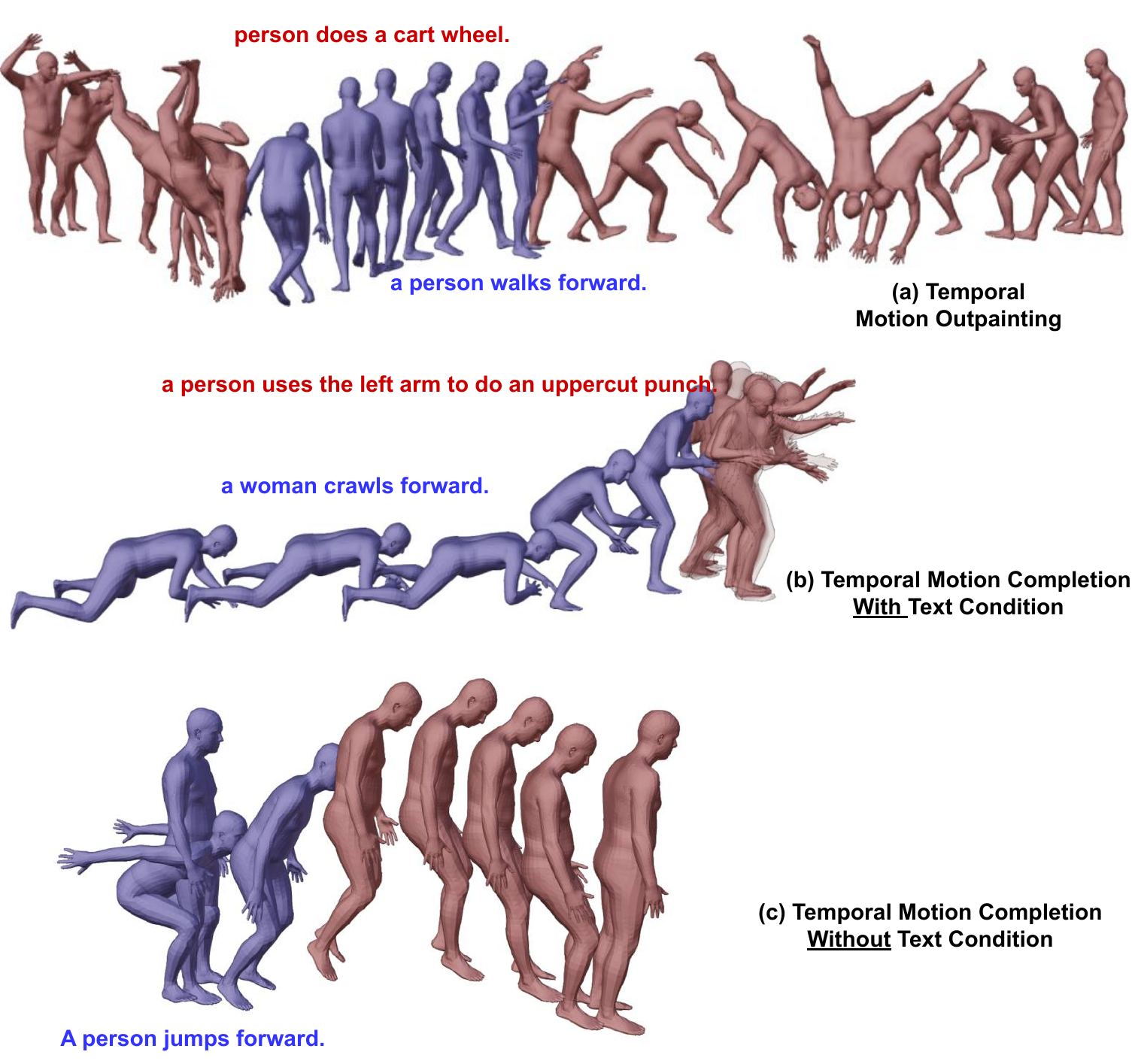}}
}
\caption{Qualitative results of (a) temporal motion outpainting and (b) temporal motion completion with text condition (c) temporal motion completion without text condition where \textcolor{red}{the red frames indicate generated motion} and \textcolor{blue}{the blue frames represent conditioned frames}.}
\label{fig:outpainting}
\end{figure}

In this section, we show that our method is not only able to edit the motion in-betweening task but also able to modify various temporal motion tasks such as temporal motion outpainting, temporal motion completion with text condition, and temporal motion completion without text condition, as shown in Figure \ref{fig:outpainting}.

\subsection{Upper Body Editing} 
\begin{figure}[H] 
\centering
\scalebox{.7}{
\centerline{\includegraphics[width=\textwidth]{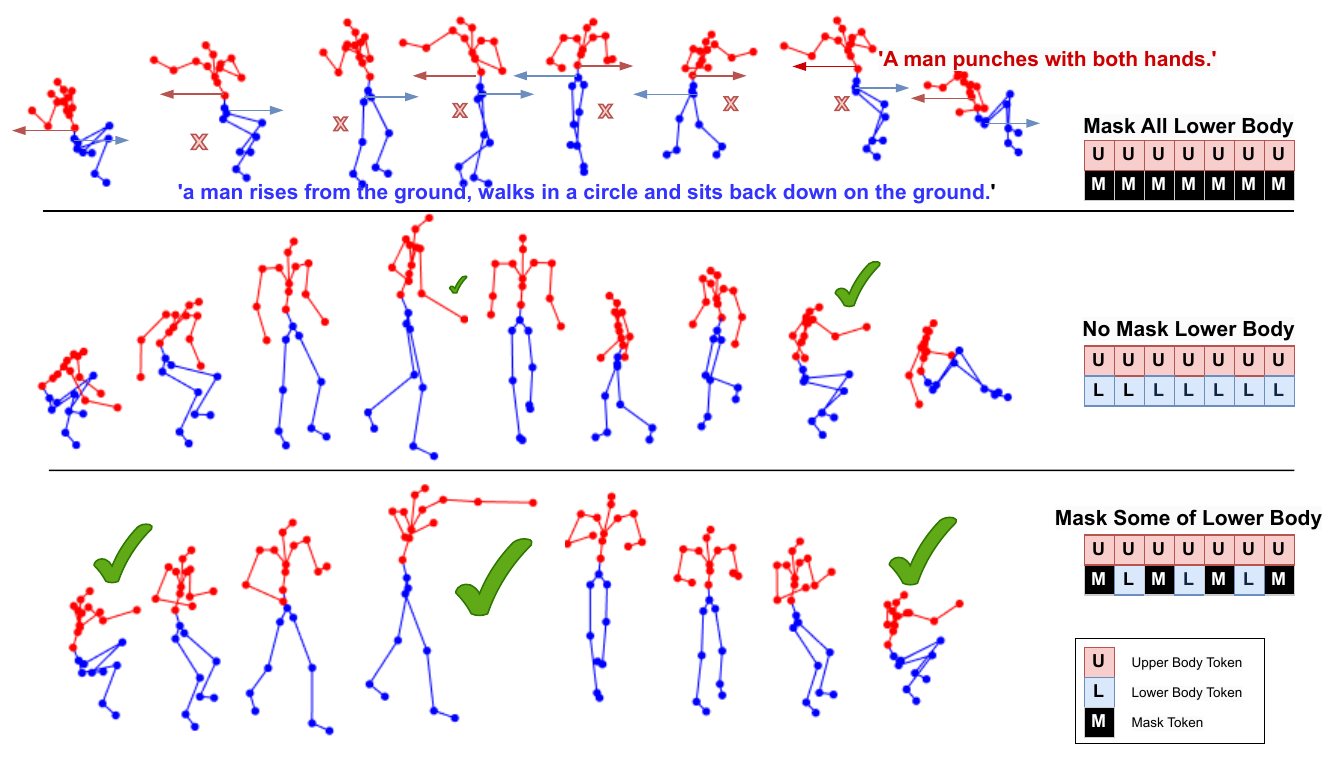}}
}
\caption{Effect of lower body masked tokens. ``\textcolor{green}{\cmark}" shows frames that exhibit strong correlation to the textual description of the upper body part}
\label{fig:lower_body_mask}
\end{figure}

The upper body editing task combines upper and lower body parts from different prompts which can lead to the out-of-distribution issue as the joint distribution of upper and lower body parts from the prompts is not present in the dataset. We overcome this challenge by introducing \texttt{[MASK]} tokens into the lower body condition tokens as controllable parameters to adjust the influence of lower body parts of the generation. As illustrated in Figure \ref{fig:lower_body_mask}, upper body parts are generated with text \textcolor{red}{``A man punches with both hands."} conditioned by lower body parts \textcolor{blue}{``a man rises from the ground, walks in a circle and sits back down on the ground."}, showing in three cases. \textbf{Top figure} shows generated upper body parts with all masked lower body tokens which means no lower body part condition at all. The result shows uncorrelated motion between upper and lower body parts. On the other hand, \textbf{middle figure} presents the generated upper body parts with all lower part conditions without any mask. The generated motion is realistic, however, less expressive as the influence from the lower body part is too strong and the joint distribution of upper and lower body parts is unseen in the training set. To solve this problem, \textbf{the bottom figure} generates motion by conditioning only some lower body tokens. This can be achieved by applying \texttt{[MASK]} tokens to some of the lower body tokens. The result shows that the motion is realistic while expressing very strong corresponding motion to both upper and lower body parts.

\subsection{Long Sequence Generation} 

\begin{figure}[H] 
\centering
\scalebox{1}{
\centerline{\includegraphics[width=\textwidth]{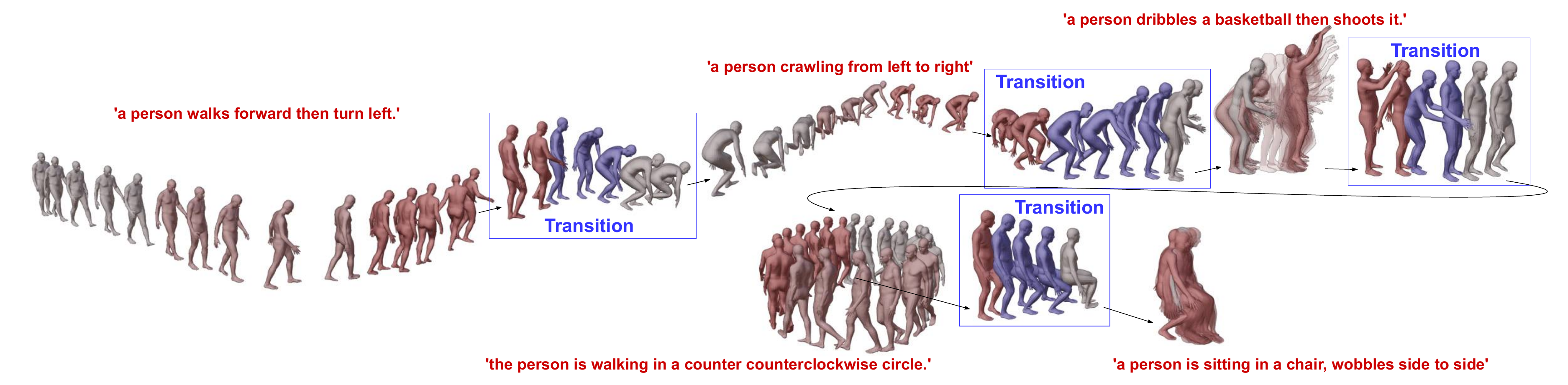}}
}
\caption{Long Sequence Generation. An arbitrary long motion sequence is generated from multiple \textcolor{red}{prompts (red frames)} combined with \textcolor{blue}{transitions (blue frames)}}
\label{fig:longrange_full}
\end{figure}

In Figure \ref{fig:longrange_full}, we generate a long motion sequence by combining multiple text prompts. First, our model generates the \textcolor{red}{motion token sequence for each prompt (red frames)}. Then, we generate \textcolor{blue}{transition motion tokens (blue frames)} conditioned on the end of the previous motion sequence and the start of the next motion sequence.

\section{Limitations}\label{sec:limitation}
Our model may face challenges in rendering some fine-grain details for exceptionally long single textual descriptions. This is due to limitations in text-to-motion training datasets, which support motions up to a maximum of 196 frames. To address this, we are exploring how our model's long motion generation capabilities can be leveraged. Specifically, we aim to integrate large language models to effectively segment a lengthy texture description into several concise text prompts. Additionally, our current model does not support the generation of interactive motions involving multiple individuals. This limitation is not unique to our model but is also present in other competing methods.  

\end{document}